\documentclass[final]{article} % For LaTeX2e
\usepackage{iclr2026_conference,times}
\usepackage{url}
\usepackage{enumitem}
\usepackage{bbm}
\usepackage{lipsum}
\usepackage{graphicx}
\usepackage{tcolorbox}
\usepackage{soul}
\usepackage{caption}
\usepackage{subcaption}
\usepackage{wrapfig}
\usepackage{lipsum}
\usepackage{enumitem}
\usepackage{multirow}
\usepackage{tikz}
\usetikzlibrary{arrows.meta}
\usepackage{basic_commands}
\usepackage[endLComment=,italicComments=false]{algpseudocodex}
\usepackage{algorithm}
\usepackage{booktabs}
\usepackage{tabularew}
\usepackage{makecell}

\AddToHook{cmd/ttfamily/after}{\frenchspacing}

\definecolor{myred}{HTML}{F54254}
\definecolor{myorange}{HTML}{FFB135}
\definecolor{mygreen}{HTML}{10BD35}
\definecolor{myblue}{HTML}{598BE7}
\definecolor{mypurple}{HTML}{9A1C6B}
\definecolor{plgray}{HTML}{999999}
\newcolumntype{R}[1]{>{\raggedleft\arraybackslash}p{#1}}
\newcolumntype{L}[1]{>{\raggedright\arraybackslash}p{#1}}
\renewcommand{\tt}[1]{\texttt{#1}}
\renewcommand{\phi}{\varphi}

\newcommand{\pl}[1]{{\color{plgray} #1}}
\newcommand{\blue}{\textcolor{myblue}}
\newcommand{\methodname}{MAC\xspace}
\newcommand{\achunk}{n}

\setlist[itemize]{itemsep=1pt, leftmargin=20pt}

\hypersetup{urlcolor=myblue,citecolor=myblue,linkcolor=myblue}

\title{Scalable Offline Model-Based RL with \\ Action Chunks}

\author{Kwanyoung Park$^1$ \quad Seohong Park$^1$ \quad Youngwoon Lee$^2$ \quad Sergey Levine$^1$ \\
    UC Berkeley$^1$ \quad Yonsei University$^2$ \\
    \url{https://kwanyoungpark.github.io/MAC/}
}

\iclrfinalcopy % Uncomment for camera-ready version, but NOT for submission.
\begin{document}

\maketitle

\begin{abstract}
In this paper, we study whether model-based reinforcement learning (RL), in particular model-based value expansion,
can provide a scalable recipe for tackling complex, long-horizon tasks in offline RL. 
Model-based value expansion fits an on-policy value function using length-$n$ imaginary rollouts generated by the current policy and a learned dynamics model.
While larger $n$ reduces bias in value bootstrapping, it amplifies accumulated model errors over long horizons, degrading future predictions.
We address this trade-off with 
an \emph{action-chunk} model that predicts a future state from a sequence of actions (an ``action chunk'')
instead of a single action, which reduces compounding errors.
In addition, instead of directly training a policy to maximize rewards,
we employ rejection sampling from an expressive behavioral action-chunk policy,
which prevents model exploitation from out-of-distribution actions.
We call this recipe \textbf{Model-Based RL with Action Chunks (\methodname)}.
Through experiments on highly challenging tasks with large-scale datasets of up to $100$M transitions,
we show that \methodname achieves the best performance among offline model-based RL algorithms,
especially on challenging long-horizon tasks.
\end{abstract}

\begin{figure}[h!]
    \centering
    \includegraphics[width=1.0\textwidth]{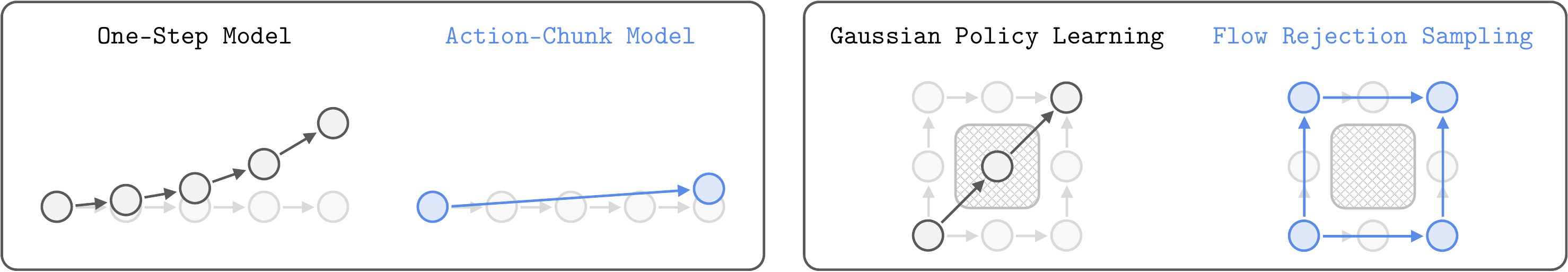}
    \caption{
    \footnotesize
    \textbf{Two main components of \methodname.}
    (\emph{Left}) Action-chunk models predict a future state given a \emph{sequence} of actions (an ``action chunk''), reducing compounding errors and enabling long-horizon model rollouts.
    (\emph{Right}) Rejection sampling from an expressive (flow) behavioral action-chunk policy enables modeling multi-modal action distributions, while preventing model exploitation from out-of-distribution actions.
    }
    \label{fig:teaser}
\end{figure}

\section{Introduction}
\label{sec:intro}

Offline reinforcement learning (RL) holds the promise of training effective decision-making agents from data, leveraging large-scale datasets.
While offline RL has achieved successes in diverse domains~\citep{scaledql_kumar2023, pac_springenberg2024},
its ability to handle complex, long-horizon tasks remains an open question.
Prior work has shown that standard, \emph{model-free} offline RL often struggles to scale to such tasks~\citep{sharsa_park2025},
hypothesizing that the cause lies in the pathologies of off-policy, temporal difference (TD) value learning.

In this work, we investigate whether an alternative paradigm,
namely \emph{model-based} RL, and in particular model-based value expansion~\citep{mve_feinberg2018},
provides a more effective recipe for long-horizon offline RL.
In this recipe, we first train a dynamics model,
and fit an \emph{on-policy} value function by rolling out the current policy within the learned model,
which is then used to update the policy.
Since on-policy value learning has demonstrated promising scalability to long-horizon tasks~\citep{dota2_berner2019, r1_guo2025},
in contrast to the relatively limited evidence for off-policy TD learning~\citep{sharsa_park2025},
we hypothesize that the combination of on-policy value learning and dynamics modeling
may also exhibit strong horizon scalability.

However, there is a tricky trade-off in this recipe.
In model-based value expansion,
we typically train a value function by rolling out the policy for $n$ steps within the model
and regressing toward the target: $V(s_t) \gets \sum_{i=0}^{n-1} \gamma^i r_{t+i} + \gamma^n \bar V(s_{t+n})$.
Here is the dilemma.
On the one hand, we want to use a large $n$ in this value update,
as this reduces the \emph{bias} in the bootstrapped target value, $\gamma^n \bar V(s_{t+n})$.
This is particularly important given that bias accumulation
is one of the major factors that hinder the scaling of offline RL~\citep{sharsa_park2025}.
On the other hand, we want to keep $n$ small enough,
as errors in the dynamics model accumulate through autoregressive queries over the horizon.
Is there a solution to this trade-off that enables long-horizon model rollouts while preventing error accumulation?

Our main hypothesis in this work is that \emph{action-chunk} models and policies,
combined with recent innovations in expressive generative models,
can provide a natural solution to the above dilemma,
enabling scaling of offline model-based RL to long-horizon tasks.
Namely, instead of training a single-step model $p(\pl{s_{t+1}} \mid \pl{s_t}, \pl{a_t})$,
we train a multi-step model $p(\pl{s_{t+n}} \mid \pl{s_t}, \pl{a_{t:t+n-1}})$ that takes an action-chunk $a_{t:t+n-1}$ as input
and predicts a future state that is $n$-step ahead.
This substantially reduces the number of recursive model calls and mitigates compounding errors (\Cref{fig:teaser}, left),
enabling long-horizon imaginary rollouts over $100$ environment steps.

To use an action-chunk model in the model-based actor-critic framework,
we need an action-chunk policy.
However, directly training a reward-maximizing action-chunk policy is challenging in offline RL,
due to the potentially multi-modal, high-dimensional action-chunk distributions in the dataset~\citep{qc_li2025}.
Hence, we employ \emph{rejection sampling} based on samples from an expressive behavioral action-chunk policy
trained with flow matching~\citep{flow_lipman2024}.
By simply defining the policy as the behavioral action-chunk sample that maximizes the value function,
we can not only capture complex action distributions from the dataset (\Cref{fig:teaser}, right),
but also effectively prevent model exploitation~\citep{morel_kidambi2020}.

We call this recipe \textbf{Model-Based RL with Action Chunks (\methodname)},
which constitutes the main contribution of this work.
Experimentally, we show that \methodname vastly improves the horizon scalability of offline model-based RL.
In particular, we demonstrate that our scalable model-based RL recipe can consume $100$M-scale data
to achieve state-of-the-art performance on highly complex, long-horizon robotic manipulation tasks from OGBench~\citep{ogbench_park2025},
often outperforming previous model-free and model-based approaches.

\section{Related Work}

\textbf{Offline model-free RL.}
Offline RL aims to learn a return-maximizing policy from a previously collected dataset,
without interaction with the environment~\citep{batch_lange2012, offline_levine2020}.
As in online RL, offline RL methods can be categorized into model-free and model-based ones.
Offline model-free RL methods train a policy without learning a dynamics model.
Prior works have proposed a number of model-free approaches based on diverse techniques,
such as
conservatism~\citep{cql_kumar2020},
behavioral regularization~\citep{brac_wu2019, awr_peng2019, awac_nair2020, td3bc_fujimoto2021, rebrac_tarasov2023, fql_park2025},
uncertainty estimation~\citep{edac_an2021, sacrnd_nikulin2023},
in-sample maximization~\citep{iql_kostrikov2022, sql_xu2023, xql_garg2023},
rejection sampling~\citep{sfbc_chen2023, idql_hansenestruch2023},
and more~\citep{onestep_brandfonbrener2021, dualrl_sikchi2024}.

\textbf{Offline model-based RL.}
In this work, we focus on offline model-based RL,
a paradigm that first trains a dynamics or trajectory model,
and then trains a policy based on rollouts generated from the learned model.
A line of work trains generative models (\eg, Transformers~\citep{transformer_vaswani2017} and diffusion models~\citep{diffusion_sohl2015, ddpm_ho2020})
to model the entire trajectory distribution of the dataset,
and typically use conditioning and guidance
to compute actions~\citep{dt_chen2021, tt_janner2021, diffuser_janner2022, mgdt_lee2022, dd_ajay2023, tap_jiang2023, hdmi_li2023, hd_chen2024, dwm_ding2024, pgd_jackson2024, jowa_cheng2025}.
Another line of work trains a (typically single-step) dynamics model,
and trains a policy based on rollouts autoregressively sampled from the learned model.
These approaches employ the learned dynamics model
for (1) ``Dyna''-style data augmentation~\citep{dyna_sutton1991, mbpo_janner2019, mopo_yu2020, morel_kidambi2020, combo_yu2021, rambo_rigter2022, mobile_sun2023, ravl_sims2024, synther_lu2023},
(2) planning~\citep{mpc_testud1978, mbop_argenson2021, iqltdmpc_chitnis2024, dmpc_zhou2025},
and (3) value estimation~\citep{mve_feinberg2018, cbop_jeong2023, leq_park2025, dreamerv3_hafner2025},
with diverse techniques to prevent model exploitation and distributional shift, such as ensemble-based uncertainty estimation.
Our method is based on model-based value expansion and falls in the third category.
However, unlike most of the previous works in this category,
we employ an \emph{action-chunk} model instead of a single-step dynamics model to reduce effective horizons and thus error accumulation.

\textbf{Horizon reduction and model-based RL.}
The curse of horizon is a fundamental challenge in reinforcement learning~\citep{horizon_liu2018, sharsa_park2025}.
In the context of model-free RL,
previous studies have proposed diverse techniques to reduce effective horizon lengths,
such as $n$-step returns to reduce the number of Bellman updates~\citep{rl_sutton2005},
and hierarchical policies to reduce the length of the effective policy horizon~\citep{hiro_nachum2018, hiql_park2023}.
Long horizons are a central challenge in model-based RL too,
since model rollouts suffer from compounding errors as the horizon grows.
Prior works in model-based RL address this challenge with
trajectory modeling~\citep{tt_janner2021, diffuser_janner2022},
hierarchical planning~\citep{hdmi_li2023, hd_chen2024},
skill-based action abstraction~\citep{skimo_shi2022},
and action-chunk multi-step dynamics modeling~\citep{m3_asadi2019, traj_lambert2021, dydiff_zhao2024, dmpc_zhou2025}.
Our work is closest to prior works that use action-chunk dynamics models.
However, these works either use the action-chunk model only for planning without having the full actor-critic loop~\citep{m3_asadi2019, traj_lambert2021, dmpc_zhou2025},
or model the entire state-action chunks~\citep{dydiff_zhao2024}.
Unlike these prior works,
we perform \emph{on-policy value learning} with an action-chunk model and policy,
while not involving additional planning or full trajectory generation.

\section{Preliminaries}

\textbf{Problem setting.}
We consider a Markov decision process (MDP) defined as
$\gM = (\gS, \gA, r, \mu, p)$,
where $\gS$ is the state space, $\gA = \sR^d$ is the action space,
$r(\pl{s}, \pl{a}): \gS \times \gA \to \sR$ is the reward function,
$\mu(\pl{s}) \in \Delta(\gS)$ is the initial state distribution,
and $p(\pl{s'} \mid \pl{s}, \pl{a}): \gS \times \gA \to \Delta(\gS)$ is the transition dynamics kernel.
$\Delta(\gX)$ denotes the set of probability distributions on a space $\gX$,
and we denote placeholder variables in \pl{gray}.
For a policy $\pi(\pl{a} \mid \pl{s}): \gS \to \Delta(\gA)$,
we define $V^\pi(s)=\E_{\tau \sim p^\pi(\pl{\tau} \mid s_0 = s)}[\sum_{t=0}^\infty \gamma^t r(s_t, a_t)]$
and $Q^\pi(s, a)=\E_{\tau \sim p^\pi(\pl{\tau} \mid s_0 = s, a_0 = a)}[\sum_{t=0}^\infty \gamma^t r(s_t, a_t)]$,
where $\gamma \in (0, 1)$ denotes the discount factor,
$\tau = (s_0, a_0, r_0, s_1, \ldots)$ denotes a trajectory,
and $p^\pi$ denotes the trajectory distribution induced by $\mu$, $p$, and $\pi$.
The goal of offline RL is to find a policy $\pi$ that maximizes $\E_{s_0 \sim \mu(\pl{s_0})}[V^\pi(s_0)]$
from an offline dataset $\gD = \{\tau^{(i)}\}$ consisting of previously collected trajectories,
with no environment interactions.

\textbf{Flow matching.}
Flow matching~\citep{flow_lipman2023, flow_albergo2023, flow_liu2023} is a technique in generative modeling
to train a velocity field whose flow generates a target distribution of interest.
As with diffusion models~\citep{diffusion_sohl2015, ddpm_ho2020},
flow models iteratively transform a noise distribution to the target distribution,
and have been shown to be highly expressive and scalable~\citep{sd3_esser2024, flow_lipman2024}.

Formally, assume that we are given a target distribution $p(\pl{x}) \in \Delta(\sR^k)$.
For a time-dependent velocity field $v(\pl{u}, \pl{x}): [0, 1] \times \sR^k \to \sR^k$
(we use $u$ to denote times in flow matching to avoid notational conflicts with environment steps in MDPs),
we define its flow, $\psi(\pl{u}, \pl{x}): [0, 1] \times \sR^k \to \sR^k$,
as the unique solution to the following ordinary differential equation (ODE)~\citep{smooth_lee2002}:
\begin{align}
    \frac{\de}{\de u} \psi(u, x) = v(u, \psi(u, x)).
\end{align}
Flow matching aims to find a velocity field whose flow
transforms a noise distribution (\eg, $k$-dimensional standard Gaussian, $\gN(0, I_d)$) at $u = 0$
to the target distribution at $u = 1$.

Prior work~\citep{flow_lipman2023, flow_albergo2023, flow_liu2023} has shown that we can train such a velocity field
by minimizing the following loss:
\begin{align}
    \E_{\substack{x_0 \sim \gN(0, I_d), \ x_1 \sim p(x), \\ u \sim \mathrm{Unif}([0, 1]), \ x_u = (1-u)x_0 + ux_1}}
    \left[ \|v(u, x_u) - (x_1 - x_0)\|_2^2 \right].
    \label{eq:flow}
\end{align}
We refer to the tutorial by \citet{flow_lipman2024} for detailed explanations and proofs.
After training the velocity field, we can obtain samples from the target distribution
by numerically following the velocity field to solve the ODE in practice
(\eg, with the Euler method).

\section{Offline Model-Based RL with Action Chunks}

\textbf{Motivation.}
Our high-level goal is to scale up offline model-based RL to complex, long-horizon decision-making problems.
Among model-based RL frameworks,
we specifically focus on model-based value expansion~\citep{mve_feinberg2018},
which combines dynamics modeling and \emph{on-policy} value learning.
This is because each of these components, namely generative modeling and on-policy RL,
has individually been shown to scale to long-horizon tasks~\citep{dota2_berner2019, fdm_harvey2022, r1_guo2025}.

In model-based value expansion, we first train a dynamics model,
and train an on-policy value function with the following update:
\begin{align}
    V(\hat s_t) \gets \sum_{i=0}^{n-1} \gamma^i r(\hat s_{t+i}, \hat a_{t+i}) + \gamma^n \bar V (\hat s_{t+n}), \label{eq:mve}
\end{align}
where $(s_t = \hat s_t, \hat a_t, \hat s_{t+1}, \ldots, \hat s_{t+n})$ is a length-$n$ imaginary rollout sampled from the model using the current policy,
and $\bar V$ is a target value function.
The policy is then updated to maximize the learned value function, and we repeat this procedure.

The problem is: how long should model rollouts be?
Unfortunately, we have two seemingly contradictory desiderata.

On the one hand, we want model rollouts to be \emph{long} enough.
If $n$ is too small,
we end up with a large number of \emph{biased} value updates with short-horizon bootstrapping in \Cref{eq:mve}.
This causes the biases to accumulate over the horizon,
which is known to be one of the main obstacles hindering value-based RL from scaling to long-horizon tasks~\citep{sharsa_park2025}.
Hence, we want to keep $n$ large enough.

On the other hand, we want model rollouts to be \emph{short} enough.
If we use a standard policy $\pi(\pl{a} \mid \pl{s})$,
we need to autoregressively call a learned dynamics model $n$ times to generate 
a length-$n$ model rollout $(\hat s_t, \hat a_t, \hat s_{t+1}, \ldots, \hat s_{t+n})$.
This makes errors in the dynamics model accumulate \emph{within} the trajectory chunk,
which would degrade performance.
Hence, we want to keep $n$ small enough.

Is there a way to naturally resolve this dilemma?

\subsection{The Idea}
\label{sec:idea}

Our main idea in this work is that a combination of an \emph{action-chunk} policy and an \emph{action-chunk} model can provide
a clean solution to the above dilemma, enabling scaling to complex, long-horizon tasks.
Specifically, we train an action-chunk model $p(\pl{s_{t+\achunk}} \mid \pl{s_t}, \pl{a_{t:t+\achunk-1}}): \gS \times \gA^\achunk \to \Delta(\gS)$
and an action-chunk policy $\pi(\pl{a_{t:t+\achunk-1}} \mid \pl{s_t}): \gS \to \Delta(\gA^\achunk)$,
where $a_{i: j}$ denotes the action chunk $(a_i, a_{i+1}, \ldots, a_{j})$.
Since each individual call of the model generates $\achunk$ actions at once,
we can reduce the number of recursive model calls by a factor of $\achunk$.
This way, we can mitigate both bias accumulation in value learning and error accumulation in model rollouts.

However, several challenges remain in implementing this idea in practice.
First, Gaussian policies, used in many previous works in offline model-based RL~\citep{mopo_yu2020, mobile_sun2023, vd4rl_lu2023, iqltdmpc_chitnis2024, leq_park2025},
are generally not expressive enough to model complex, multi-modal action-\emph{chunk} distributions (\Cref{fig:teaser}).
Second, penalizing out-of-distribution actions based on uncertainty in the dynamics model, as typically done by prior work in offline model-based RL~\citep{mopo_yu2020, morel_kidambi2020, mobile_sun2023},
can be challenging due to the potentially high complexity of the action-chunked dynamics distribution.

To handle these challenges,
we employ rejection sampling from an expressive behavioral action-chunk policy.
Specifically, we use flow matching~\citep{flow_lipman2024} to train a behavioral cloning (BC) action-chunk policy,
and \emph{define} a policy as the $\argmax$ action chunk
(among $N$ chunks sampled from the BC policy) that maximizes the learned value function:
\begin{align}
    \pi(s_t) :\stackrel{d}{=} \argmax_{\{a_{t:t+\achunk-1}^{(i)}\}_{i=1}^N \sim \pi^\beta(\pl{a_{t:t+\achunk-1}} \mid s_t)} Q(s_t, a^{(i)}_{t:t+\achunk-1}),
    \label{eq:rej}
\end{align}
where $\pi^\beta(\pl{a_{t:t+\achunk-1}} \mid \pl{s_t}): \gS \to \Delta(\gA^\achunk)$ denotes an action-chunk flow BC policy,
$Q(\pl{s_t}, \pl{a_{t:t+\achunk-1}}): \gS \times \gA^\achunk \to \sR$ denotes an action-chunk value function,
and $\stackrel{d}{=}$ denotes equality in distribution.

Thanks to the expressivity of the flow BC policy,
we query the model only with in-distribution action-chunk samples,
which obviates the need for an additional uncertainty penalization mechanism.
Moreover, rejection sampling is generally more robust to hyperparameters~\citep{dmpc_zhou2025, sharsa_park2025},
making our method simpler and easier to tune than other alternatives,
which may require tuning an uncertainty penalization coefficient for each task.

\begin{figure}[t]
\vspace{-1.5em}
\begin{algorithm}[H]
\caption{Offline Model-Based RL with Action Chunks (\methodname)}
\label{alg:training}
\begin{algorithmic}
    \Require Dataset $\mathcal{D}$, rollout length $H$, action chunking size $n$, rejection sampling size $M$ 

    \State
    \State \blue{\emph{// Training loop}}
    \While{not converged}
        \State \Comment{Sample action-chunked batch from the dataset ($\bm{a}_t = a_{t:t+n-1}$, $\bm{r}_t = \sum_{i=0}^{\achunk-1} \gamma^i r_{t+i}$)}
        \State Sample batch $\{(s_t, \bm{a}_{t}, \bm{r}_{t},  s_{t+\achunk})\} \sim \mathcal{D}$     
        
        \State
        \State \Comment{Train \textbf{BC policy} using dataset transitions}
        \State Update flow BC policy $\pi_\theta$ with flow-matching loss (\Cref{eq:flow_ac})
        \State Update one-step BC policy $\pi_\omega$ with distillation loss (\Cref{eq:distill_ac})

        \State
        \State \Comment{Train \textbf{dynamics} and \textbf{reward} model using dataset transitions}
        \State Update dynamics model $p_\psi$ to minimize 
        $\E [\|p_\psi(s_t, \bm{a}_t) - s_{t+\achunk}\|_2^2]$ (\Cref{eq:dyn}) 
        \State Update reward model $r_\psi$ to minimize 
        $\E [\left\|r_\psi(s_t, \bm{a}_t) - \bm{r}_t \right\|_2^2]$ (\Cref{eq:rew}) 

        \vspace{0.2em}
        \State
        \State \Comment{Generate \textbf{model rollouts} ($\hat{s}_t = s_t$)}
        \For{$k=0, 1, \ldots, H-1$}
            \State $\bm{\hat{a}}_{t+kn} \leftarrow ~$\textsc{policy}$(\hat{s}_{t+kn})$
            \State $\hat{s}_{t+(k+1)\achunk}, \bm{\hat{r}}_{t+kn} \sim p_{\psi}(\cdot \mid \hat{s}_{t+kn}, \bm{\hat{a}}_{t+kn})$, $r_{\psi}(\cdot \mid \hat{s}_{t+kn}, \bm{\hat{a}}_{t+kn})$
        \EndFor

        \State
        \State \Comment{Update \textbf{value} using model rollouts}
        \State Update value $V_\phi$ with $nH$-step targets from the rollout (\Cref{eq:loss_v})

        \State
        \State \Comment{Learn \textbf{critic} for faster rejection sampling}
        \State Update critic $Q_\phi$ with the learned value function $V_{\phi}$ (\Cref{eq:loss_q})
    \EndWhile

    \State
    \State \blue{\emph{// Extract action from flow BC policy $\pi_{\theta}$ with rejection sampling}}
    \Function{policy}{$s$}
        \State $z \sim \mathcal{N}(0, I)$
        \State $\bm{\hat{a}}^{(i)} = \pi_{\omega}(s, z)$
        
        \Return $\text{argmax}_{\bm{\hat{a}}^{(1)}, \cdots, \bm{\hat{a}}^{(M)}} Q_{\phi} (s, \bm{\hat{a}}^{(i)})$
    \EndFunction
\end{algorithmic}
\end{algorithm}
\vspace{-2em}
\end{figure}

\subsection{Practical Algorithm}

Based on the idea discussed in the previous section,
we now describe the full details of our method for scalable offline model-based RL,
which we call \textbf{Model-Based RL with Action Chunks (\methodname)}.
\methodname consists of the following components:
an action-chunk dynamics model $p_\psi$,
an action-chunk reward model $r_\psi$,
a flow action-chunk policy $\pi_\theta$ 
, and value functions $V_\phi$ and $Q_\phi$. 
For notational simplicity, we override the symbols $\psi$, and $\phi$ to denote all model-, and value-related parameters, respectively. Moreover, we denote $\bm{a}_t \in \gA^{\achunk}$ to be the action chunk $a_{t:t+\achunk}$, and $\bm{r}_t$ to be the sum of discounted rewards for $n$ steps $\sum_{i=0}^{n-1} \gamma^i r_{t+i}$.

\textbf{Action-chunk dynamics and reward models.}
% The training objective for the action-chunk model is simple.
For dynamics modeling,
we minimize the following losses to train a deterministic action-chunk dynamics model
$p_\psi(\pl{s_t}, \pl{\bm{a}_{t}}): \gS \times \gA^\achunk \to \gS$
and an action-chunk reward model
$r_\psi(\pl{s_t}, \pl{\bm{a}_{t}}): \gS \times \gA^\achunk \to \sR$:
\begin{align}
    L^\mathrm{dyn}(\psi) &= \E_{(s_t, a_t, \cdots, s_{t+\achunk}) \sim \gD} \left[\|p_\psi(s_t, \bm{a}_t) - s_{t+\achunk}\|_2^2\right], \label{eq:dyn} \\
    L^\mathrm{rew}(\psi) &= \E_{(s_t, a_t, \cdots, s_{t+n}) \sim \gD} \left[\left\|r_\psi(s_t, \bm{a}_t) - \bm{r}_t \right\|_2^2\right], \label{eq:rew}
\end{align}
where trajectory chunks are uniformly sampled from the offline dataset.
The dynamics function $p_\psi$ is modeled by a deterministic multi-layer perceptron (MLP).
While we found this to be sufficient in our benchmark environments,
we note that it is possible to replace the MLP with an expressive flow model (as in our policy)
in stochastic or partially observable environments.

\textbf{Flow action-chunk policies.}
For the actor, we employ rejection sampling using a behavioral flow action-chunk policy, as described in \Cref{sec:idea}.
To train a flow BC policy, we train a state-dependent velocity field
$v_\theta: \sR \times \gS \times \gA^n \to \gA^n$,
with the flow-matching loss (\Cref{eq:flow}):
\begin{align}
    L^\mathrm{flow}(\theta) &= \E_{\substack{z \sim \gN(0, I_{\achunk d}), \ (s_t, \bm{a}_{t}) \sim \gD, \\ u \sim \mathrm{Unif}([0, 1]), \ a_z = (1-u)z+uz}}\left[\| v_\theta(u, s_t, a_z) - (\bm{a}_t - z) \|_2^2\right].
    \label{eq:flow_ac}
\end{align}
We define $\pi_\theta(s_t, z) \in \gA^n$ as the
destination of the induced flow at $u=1$
when starting with $(s_t, z)$ at $u=0$ and following the velocity field $v_\theta$.
Then, by sampling multiple noises $z \sim \gN(0, I_{nd})$ and computing $\pi_\theta(s_t, z)$,
we can obtain behavioral action-chunk samples,
which are then used for rejection sampling (\Cref{eq:rej}) along with a learned value function (described in the ``Value learning'' section below).

One issue with this rejection sampling framework is speed.
To compute a single action chunk using \Cref{eq:rej},
we need $NF$ queries of the velocity field $v_\theta$,
where $N$ is the number of samples and $F$ is the number of flow steps in the Euler method.
For example, with $N=8$ and $F=10$,
we need to query the velocity field $80$ times to sample a single action chunk.
This is particularly prohibitive in model-based RL,
as we need to sample multiple imaginary rollouts during training in batches,
unlike methods that employ rejection sampling only at test time~\citep{idql_hansenestruch2023, sharsa_park2025, dmpc_zhou2025}.

To address this issue, we train an additional \emph{one-step} flow policy
that directly predicts the output of the ODE flow policy.
Specifically, we train a one-step MLP action-chunk policy
$\pi_\omega(\pl{s_t}, \pl{z}): \gS \times \gA^n \to \gA^n$ parameterized by $\omega$,
with the following flow distillation loss~\citep{fql_park2025}:
\begin{align}
    L^\mathrm{distill}(\omega) &= \E_{s_t \sim \gD, \ z \sim \gN(0, I_{ d\achunk})}\left[\| \pi_\omega(s_t, z) - [\pi_\theta(s_t, z)]_\times \|_2^2\right],
    \label{eq:distill_ac}
\end{align}
where $[ \cdot ]_\times$ denotes the ``stop gradient'' operation.

Unlike the ODE policy, $\pi_\omega$ only requires a single network call to produce an action chunk,
reducing the number of queries from $NF$ to $N$ for rejection sampling in \Cref{eq:rej}.
This substantially reduces both the training and inference cost of \methodname.

\textbf{Value learning.}
In \methodname, value functions are trained from on-policy model rollouts (i.e., imaginary trajectories).
To train value functions,
we first generate $M$ imaginary (action-chunk) trajectories of length $H$,
\begin{align}
    \gD^\mathrm{img} = \{(s_t^{(i)}, \bm{\hat{a}}_{t}^{(i)}, \bm{\hat r}_{t}^{(i)}, \hat s_{t+\achunk}^{(i)}, \bm{\hat{a}}_{t+\achunk}^{(i)}, \bm{\hat r}_{t+\achunk}^{(i)}, \ldots, \hat s_{t+H\achunk}^{(i)})\}_{i=1}^M,
\end{align}
where $\bm{\hat{a}}_{t}$ denotes the action chunk $\hat{a}_{t:t+n}$ generated from the rejection sampling policy, and $\bm{\hat{r}}_{t}$ denotes the discounted sum of rewards $\sum_{i=0}^{n-1} \gamma^i r_{t+i}$ predicted from the reward model $r_{\psi}(\cdot | s_t, \bm{a}_t)$.
Here, initial states $s_t^{(i)}$ are uniformly sampled from the dataset,
and subsequent actions, rewards, and next states are synthesized by our rejection-sampling policy, reward model, and dynamics model, respectively, hence the hat notation.

After collecting $\gD^\mathrm{img}$, we update the value function
$V_\phi(\pl{s_t}): \gS \to \sR$
with the following loss:
\begin{align}
L^V(\phi) &= \E\left[\left(
V_\phi(\hat s_{t+kn}) - \sum_{i=k}^{H-1} \gamma^{(i-k)n} \bm{\hat r}_{t+in} - \gamma^{(H-k)n}V_{\bar \phi}(\hat s_{t+Hn})
\right)^2\right], \label{eq:loss_v} 
\end{align}
where $\bar \phi$ denotes exponentially averaged target parameters~\citep{dqn_mnih2013},
and the expectations are over
$(s_t=\hat s_t, \bm{\hat a}_t, \bm{\hat r}_t, \ldots, \hat s_{t+Hn})$ uniformly sampled from $\gD^\mathrm{img}$
and $k$ uniformly sampled from $\{0, 1, \ldots, H-1\}$.

Finally, we train the action-chunk Q function 
$Q_\phi(\pl{s_t}, \pl{\bm{a}_{t}}): \gS \times \gA^n \to \sR$ for the rejection sampling
with the following loss:
\begin{align}
L^Q(\phi) &= \E_{s_t \sim \gD} \left[\left(
Q_\phi(s_{t}, \bm{\hat a}_{t}) - \bm{\hat r}_{t} - \gamma^n [V_\phi(\hat s_{t+n})]_\times
\right)^2\right]. \label{eq:loss_q}
\end{align}
We do not reuse $\gD^\mathrm{img}$ after performing one gradient update of value functions;
\ie, we generate new model rollouts every epoch.
We provide a pseudocode for \methodname in \Cref{alg:training}.

\textbf{Notes on hyperparameters.}
While \methodname has several learnable components,
we found that \methodname is highly robust to hyperparameters in our experiments.
In particular, we use the same horizon hyperparameters of $(n, H) = (10, 10)$
for \textbf{all} tasks considered in this work.
We also use the same number ($N = 32$) of samples for flow rejection sampling during evaluation across all tasks.
See \Cref{sec:exp_details} for the full details.

\section{Experiments}
\label{sec:exp}
Now, we empirically evaluate the performance of \methodname through a series of experiments.
Our main research question is how well \methodname scales to \emph{long-horizon} tasks
compared to previous offline model-based RL approaches, which we answer in \Cref{sec:exp_long}.
Then, we compare \methodname with previous methods on standard offline RL benchmark tasks
to assess its effectiveness as a general offline RL algorithm (\Cref{sec:exp_standard}).
Finally, we provide several analyses and ablation studies
to understand the importance of each component of \methodname (\Cref{sec:exp_qna}).
In our experiments, we use four random seeds (unless otherwise mentioned) and report standard deviations in tables and $95\%$ confidence intervals in plots.
In tables, we highlight numbers that are above or equal to $95\%$ of the best performance.

\subsection{Experiments on Large-Scale, Long-Horizon Tasks}
\label{sec:exp_long}

We first study the horizon scalability of \methodname by evaluating it on large-scale, long-horizon benchmark tasks.

\textbf{Tasks and datasets.}
To assess the scalability limits of each algorithm,
we employ three highly challenging, long-horizon simulated robotic tasks used in the work by \citet{sharsa_park2025} modified from OGBench~\citep{ogbench_park2025}:
\tt{humanoidmaze-giant}, \tt{cube-octuple}, and \tt{puzzle-4x5}.
These tasks are not just long-horizon but also goal-conditioned
(\ie, the agent must reach any goal states given at test time),
requiring complex, \emph{multi-task} reasoning over a long episode.
They present a variety of control challenges
from high-dimensional humanoid navigation
to complex object manipulation and combinatorial puzzle solving.
The hardest task in each environment requires $700$--$3000$ environment steps
and $8$--$20$ different atomic motions to complete.
In addition to these long-horizon tasks,
we also evaluate methods on shorter-horizon variants in each category
(\ie, \tt{humanoidmaze-medium}, \tt{cube-double}, and \tt{puzzle-3x3})
to examine each method's ability to handle different horizon lengths.

For datasets, we mainly employ the $100$M-transition datasets provided by \citet{sharsa_park2025}.
These large-scale datasets are collected in a task-agnostic manner (\eg, trajectories consisting of random atomic motions),
meaning that the agent must understand the dynamics and stitch different parts of trajectories to achieve test-time tasks.

\textbf{Methods.}
We mainly compare \methodname against six previous model-based RL methods across diverse categories,
including flat and hierarchical, and actor-critic and planning approaches.

Among standard model-based RL approaches,
we consider MOPO, MOBILE, LEQ, and F-MPC.
MOPO~\citep{mopo_yu2020} and MOBILE~\citep{mobile_sun2023} are Dyna-style methods
(\ie, ones that generate imaginary rollouts, augment the dataset, and run off-policy RL)
combined with uncertainty penalization, based on model uncertainty and Bellman inconsistency, respectively.
LEQ~\citep{leq_park2025} is a model-based actor-critic method
based on conservative return estimation.
F-MPC is a flow-based variant of D-MPC~\citep{dmpc_zhou2025},
which trains an action-chunk dynamics model as in our method,
but performs planning (based on a behavioral Monte Carlo value function)
instead of training an on-policy value function with actor-critic.

\clearpage
Among sequence modeling approaches,
we consider Diffuser and HD-DA.
Diffuser~\citep{diffuser_janner2022} models trajectories with diffusion~\citep{ddpm_ho2020} for planning,
and HD-DA~\citep{hd_chen2024} extends Diffuser using hierarchical  models and high-level planning
to handle long horizons.

For reference, we additionally consider three performant model-free RL algorithms as well:
IQL, $n$-SAC+BC, and SHARSA.
IQL~\citep{iql_kostrikov2022} is a standard model-free offline RL algorithm based on in-sample value learning.
$n$-SAC+BC~\citep{sharsa_park2025} is a behavior-regularized offline RL method
that employs $n$-step returns to handle long horizons.
SHARSA~\citep{sharsa_park2025} is a state-of-the-art offline RL algorithm designed for long-horizon tasks that employs hierarchical policies and flow rejection sampling.

\begin{table}[t!]
\caption{
\footnotesize
\textbf{Results on large-scale, long-horizon tasks.}
\methodname achieves the best performance among model-based RL algorithms.
}
\label{table:long_100m}
\vspace{-0.5em}

\centering
\scalebox{0.62}
{

\begin{tabularew}{l*{10}{>{\spew{.5}{+1}}r@{\,}l}}
\toprule

\multicolumn{1}{c}{} & \multicolumn{6}{c}{\tt{Model-Free}} & \multicolumn{4}{c}{\tt{Seq. Modeling}} & \multicolumn{10}{c}{\tt{Model-Based}} \\
\cmidrule(r){2-7}
\cmidrule(r){8-11}
\cmidrule(r){12-21}
\multicolumn{1}{l}{\tt{Environment}} & \multicolumn{2}{c}{\tt{GCIQL}} & \multicolumn{2}{c}{\tt{n-SAC+BC}} & \multicolumn{2}{c}{\tt{SHARSA}} & \multicolumn{2}{c}{\tt{Diffuser}} & \multicolumn{2}{c}{\tt{HD-DA}} & \multicolumn{2}{c}{\tt{MOPO}} & \multicolumn{2}{c}{\tt{MOBILE}} & \multicolumn{2}{c}{\tt{LEQ}} & \multicolumn{2}{c}{\tt{FMPC}} & \multicolumn{2}{c}{\tt{\color{myblue}\tt{MAC}}} \\
\midrule
\tt{humanoidmaze-medium-navigate-oraclerep-v0} & \tt{55} &{\tiny $\pm$\tt{1}} & \tt{\color{myblue}{98}} &{\tiny $\pm$\tt{2}} & \tt{\color{myblue}{95}} &{\tiny $\pm$\tt{2}} & \tt{0} &{\tiny $\pm$\tt{0}} & \tt{0} &{\tiny $\pm$\tt{0}} & \tt{27} &{\tiny $\pm$\tt{5}} & \tt{23} &{\tiny $\pm$\tt{3}} & \tt{0} &{\tiny $\pm$\tt{0}} & \tt{18} &{\tiny $\pm$\tt{5}} & \tt{36} &{\tiny $\pm$\tt{2}} \\
\tt{humanoidmaze-giant-navigate-oraclerep-v0} & \tt{4} &{\tiny $\pm$\tt{2}} & \tt{\color{myblue}{82}} &{\tiny $\pm$\tt{5}} & \tt{43} &{\tiny $\pm$\tt{6}} & \tt{0} &{\tiny $\pm$\tt{0}} & \tt{0} &{\tiny $\pm$\tt{0}} & \tt{0} &{\tiny $\pm$\tt{0}} & \tt{0} &{\tiny $\pm$\tt{0}} & \tt{0} &{\tiny $\pm$\tt{0}} & \tt{0} &{\tiny $\pm$\tt{0}} & \tt{0} &{\tiny $\pm$\tt{0}} \\
\tt{cube-double-play-oraclerep-v0} & \tt{74} &{\tiny $\pm$\tt{3}} & \tt{32} &{\tiny $\pm$\tt{20}} & \tt{\color{myblue}{95}} &{\tiny $\pm$\tt{3}} & \tt{1} &{\tiny $\pm$\tt{1}} & \tt{2} &{\tiny $\pm$\tt{1}} & \tt{25} &{\tiny $\pm$\tt{12}} & \tt{15} &{\tiny $\pm$\tt{3}} & \tt{0} &{\tiny $\pm$\tt{0}} & \tt{37} &{\tiny $\pm$\tt{13}} & \tt{\color{myblue}{100}} &{\tiny $\pm$\tt{1}} \\
\tt{cube-octuple-play-oraclerep-v0} & \tt{0} &{\tiny $\pm$\tt{0}} & \tt{0} &{\tiny $\pm$\tt{0}} & \tt{19} &{\tiny $\pm$\tt{3}} & \tt{0} &{\tiny $\pm$\tt{0}} & \tt{0} &{\tiny $\pm$\tt{0}} & \tt{0} &{\tiny $\pm$\tt{0}} & \tt{0} &{\tiny $\pm$\tt{0}} & \tt{0} &{\tiny $\pm$\tt{0}} & \tt{0} &{\tiny $\pm$\tt{0}} & \tt{\color{myblue}{30}} &{\tiny $\pm$\tt{6}} \\
\tt{puzzle-3x3-play-oraclerep-v0} & \tt{\color{myblue}{98}} &{\tiny $\pm$\tt{3}} & \tt{91} &{\tiny $\pm$\tt{2}} & \tt{\color{myblue}{100}} &{\tiny $\pm$\tt{0}} & \tt{1} &{\tiny $\pm$\tt{1}} & \tt{1} &{\tiny $\pm$\tt{1}} & \tt{19} &{\tiny $\pm$\tt{2}} & \tt{15} &{\tiny $\pm$\tt{5}} & \tt{1} &{\tiny $\pm$\tt{1}} & \tt{12} &{\tiny $\pm$\tt{6}} & \tt{\color{myblue}{100}} &{\tiny $\pm$\tt{0}} \\
\tt{puzzle-4x5-play-oraclerep-v0} & \tt{20} &{\tiny $\pm$\tt{1}} & \tt{19} &{\tiny $\pm$\tt{4}} & \tt{91} &{\tiny $\pm$\tt{4}} & \tt{0} &{\tiny $\pm$\tt{0}} & \tt{0} &{\tiny $\pm$\tt{0}} & \tt{0} &{\tiny $\pm$\tt{0}} & \tt{0} &{\tiny $\pm$\tt{0}} & \tt{1} &{\tiny $\pm$\tt{3}} & \tt{0} &{\tiny $\pm$\tt{0}} & \tt{\color{myblue}{99}} &{\tiny $\pm$\tt{3}} \\
\bottomrule
\end{tabularew}

}
\vspace{-0.5em}
\end{table}

\subsubsection{Results}
We present the main comparison results
on six tasks in \Cref{table:long_100m}.
The results suggest that \methodname achieves the best performance across all settings
among model-based RL algorithms.
In particular, none of the previous model-based RL approaches achieves non-trivial performance on three long-horizon tasks.
This is likely because they either use single-step models, which suffer from error accumulation (see \Cref{fig:mse}),
or are based on planning, which is insufficient to perform full-fledged long-horizon dynamic programming.
Moreover, even compared to state-of-the-art model-free RL approaches (\eg, SHARSA),
\methodname achieves the best performance on four out of six tasks,
especially on long-horizon manipulation tasks (\tt{cube-octuple} and \tt{puzzle-4x5}).

\textbf{Negative results.}
Despite its strength on manipulation tasks,
\methodname, as well as all other model-based RL approaches,
struggles on long-horizon robotic locomotion tasks (\eg, \tt{humanoidmaze-giant}).
This is a widely known phenomenon;
prior works~\citep{iqltdmpc_chitnis2024, leq_park2025} have also found that model-based RL particularly struggles
in similar robotic maze navigation environments
(\eg, \tt{antmaze-large} in D4RL~\citep{d4rl_fu2020}).
We believe this is mainly due to the difficulties in modeling \emph{contact-rich} dynamics in locomotion domains,
where dynamics tend to be highly erratic due to discontinuities, resulting in severe model error accumulation.
While \methodname's action-chunk dynamics model does mitigate this issue to some extent, leading to the best performance among model-based RL approaches (\Cref{table:long_100m}),
it is not sufficient to fully close the gap between model-free and model-based approaches on these locomotion tasks.
We believe this issue may be addressed by more expressive generative models or latent dynamics models,
which we leave for future work.

\vspace{-0.2em}
\subsection{Experiments on Standard Benchmarks}
\vspace{-0.1em}
\label{sec:exp_standard}

Next, we evaluate \methodname on standard, reward-based benchmark tasks
to assess its ability to serve as a general offline RL algorithm under limited data.

\textbf{Tasks and datasets.}
We employ $25$ single-task manipulation tasks from five environments in OGBench~\citep{ogbench_park2025}:
\tt{cube-\{single, double\}}, \tt{scene}, and \tt{puzzle-\{3x3, 4x4\}}.
Unlike in \Cref{sec:exp_long}, these tasks are reward-based (\ie, not goal-conditioned),
where the agent gets a reward according to the progress of the task. %whenever it succeeds at a subtask.
We use the $1$M-sized \tt{play} datasets given by the benchmark. We report the average success rate across $5$ tasks for each environment. 

\textbf{Methods.}
For model-based approaches, we consider the four standard model-based RL algorithms used in \Cref{sec:exp_long}.
Additionally,
we consider four standard, performant model-free RL algorithms
used in the work by \citet{fql_park2025}:
IQL~\citep{iql_kostrikov2022}, ReBRAC~\citep{rebrac_tarasov2023}, IDQL~\citep{idql_hansenestruch2023}, and FQL~\citep{fql_park2025}.
Among them, FQL is a state-of-the-art model-free offline RL method on these tasks.

\clearpage

\begin{table}[t!]
\caption{
\footnotesize
\textbf{Results on standard reward-based benchmark tasks.}
\methodname achieves the best performance among both model-based and model-free RL algorithms.
}
\label{table:standard}

\vspace{-0.5em}
\centering
\scalebox{0.62}
{

\begin{tabular*}{1.61\textwidth}{p{7.7cm} *{8}{R{0.5cm}@{\hspace{0.05cm}}L{0.33cm}}R{0.6cm}@{\hspace{0.05cm}}L{0.43cm}}
\toprule
\multicolumn{1}{c}{} & \multicolumn{8}{c}{\tt{Model-Free}} & \multicolumn{10}{c}{\tt{Model-Based}} \\
\cmidrule(r){2-9}
\cmidrule(r){10-19}
\multicolumn{1}{l}{\tt{Environment}} & \multicolumn{2}{c}{\makebox[1.15cm]{\hspace{-0.05cm}\tt{IQL}}} & \multicolumn{2}{c}{\makebox[1.15cm]{\hspace{-0.05cm}\tt{ReBRAC}}} & \multicolumn{2}{c}{\makebox[1.15cm]{\hspace{-0.05cm}\tt{IDQL}}} & \multicolumn{2}{c}{\makebox[1.15cm]{\hspace{-0.05cm}\tt{FQL}}} & \multicolumn{2}{c}{\makebox[1.15cm]{\hspace{0.05cm}\tt{MOPO}}} & \multicolumn{2}{c}{\makebox[1.15cm]{\hspace{0.02cm}\tt{MOBILE}}} & \multicolumn{2}{c}{\makebox[1.15cm]{\hspace{0.1cm}\tt{LEQ}}} & \multicolumn{2}{c}{\makebox[1.15cm]{\hspace{0.1cm}\tt{FMPC}}} & \multicolumn{2}{c}{\makebox[1.3cm]{\tt{\color{myblue}\tt{MAC}}}} \\
\midrule
\tt{cube-single-play-v0 (5 tasks)}& \tt{83} &{\tiny $\pm$\tt{9}} & \tt{91} &{\tiny $\pm$\tt{5}} & \tt{\color{myblue}{95}} &{\tiny $\pm$\tt{4}} & \tt{\color{myblue}{96}} &{\tiny $\pm$\tt{3}} & \tt{12} &{\tiny $\pm$\tt{4}} & \tt{81} &{\tiny $\pm$\tt{8}} & \tt{0} &{\tiny $\pm$\tt{0}} & \tt{9} &{\tiny $\pm$\tt{5}} & \tt{\color{myblue}{99}} &{\tiny $\pm$\tt{2}} \\
\tt{cube-double-play-v0 (5 tasks)}& \tt{7} &{\tiny $\pm$\tt{11}} & \tt{12} &{\tiny $\pm$\tt{17}} & \tt{15} &{\tiny $\pm$\tt{17}} & \tt{29} &{\tiny $\pm$\tt{21}} & \tt{1} &{\tiny $\pm$\tt{1}} & \tt{1} &{\tiny $\pm$\tt{2}} & \tt{0} &{\tiny $\pm$\tt{0}} & \tt{3} &{\tiny $\pm$\tt{2}} & \tt{\color{myblue}{53}} &{\tiny $\pm$\tt{4}} \\
\tt{scene-play-v0 (5 tasks)}& \tt{28} &{\tiny $\pm$\tt{36}} & \tt{41} &{\tiny $\pm$\tt{37}} & \tt{46} &{\tiny $\pm$\tt{44}} & \tt{56} &{\tiny $\pm$\tt{45}} & \tt{6} &{\tiny $\pm$\tt{8}} & \tt{8} &{\tiny $\pm$\tt{4}} & \tt{0} &{\tiny $\pm$\tt{0}} & \tt{4} &{\tiny $\pm$\tt{4}} & \tt{\color{myblue}{97}} &{\tiny $\pm$\tt{4}} \\
\tt{puzzle-3x3-play-v0 (5 tasks)}& \tt{9} &{\tiny $\pm$\tt{13}} & \tt{21} &{\tiny $\pm$\tt{38}} & \tt{10} &{\tiny $\pm$\tt{21}} & \tt{\color{myblue}{30}} &{\tiny $\pm$\tt{31}} & \tt{20} &{\tiny $\pm$\tt{0}} & \tt{12} &{\tiny $\pm$\tt{9}} & \tt{10} &{\tiny $\pm$\tt{7}} & \tt{1} &{\tiny $\pm$\tt{1}} & \tt{20} &{\tiny $\pm$\tt{0}} \\
\tt{puzzle-4x4-play-v0 (5 tasks)}& \tt{7} &{\tiny $\pm$\tt{4}} & \tt{14} &{\tiny $\pm$\tt{8}} & \tt{29} &{\tiny $\pm$\tt{13}} & \tt{17} &{\tiny $\pm$\tt{10}} & \tt{0} &{\tiny $\pm$\tt{0}} & \tt{0} &{\tiny $\pm$\tt{0}} & \tt{0} &{\tiny $\pm$\tt{0}} & \tt{0} &{\tiny $\pm$\tt{0}} & \tt{\color{myblue}{78}} &{\tiny $\pm$\tt{13}} \\
\bottomrule
\end{tabular*}
}
\vspace{-1.0em}
\end{table}

\subsubsection{Results}

\Cref{table:standard} summarizes the comparison results on $25$ standard benchmark tasks.
The results show that \methodname achieves the best average performance on four out of five environments.
Notably, \methodname achieves substantially better performance than all other methods especially on (relatively) long-horizon environments,
such as \tt{cube-double}, \tt{scene}, and \tt{puzzle-4x4},
showing the promise of offline model-based RL in manipulation domains.

\vspace{-0.5em}
\subsection{Q\&As}
\label{sec:exp_qna}

In this section, we discuss and analyze the components of \methodname through the following Q\&As.

\ul{\textbf{Q: Do action chunks actually mitigate error accumulation?}}

\begin{wrapfigure}{r}{0.34\textwidth}
    \vspace{-0.1em}
    \centering
    \raisebox{0pt}[\dimexpr\height-1.0\baselineskip\relax]{
        \begin{subfigure}[t]{1.0\linewidth}
        \includegraphics[width=\linewidth]{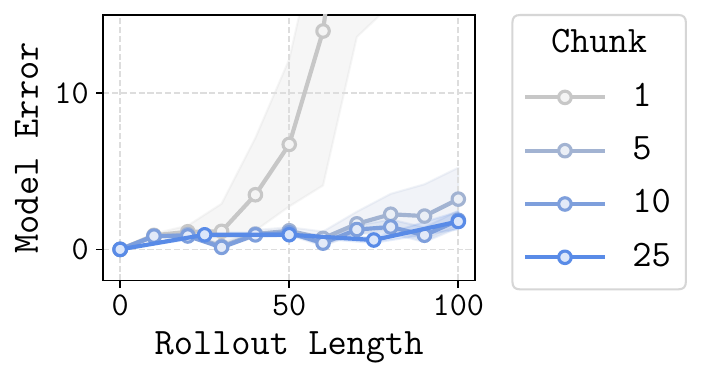}
        \end{subfigure}
    }
    \vspace{-2.0em}
    \caption{
    \footnotesize
        \textbf{Action chunking reduces model errors.}
    }
    \vspace{-1.5em}
    \label{fig:mse}
\end{wrapfigure}
\textbf{A:}
Our main motivation for using action chunking is to reduce error accumulation in autoregressive trajectory generation.
However, one might question whether it is actually the case,
given that increasing the action chunk length also increases the difficulty of learning the model.
To examine this, we analyze how the chunk length affects model errors.
Specifically, we train dynamics models with action chunk lengths of $\{1, 5, 10, 25\}$
and measure their mean squared prediction errors along a length-$100$ dataset trajectory in \tt{puzzle-4x5}.
\Cref{fig:mse} presents the result,
suggesting that longer action chunks indeed substantially mitigate error accumulation.
Notably, the errors from a standard one-step model diverge over time,
substantiating the necessity of multi-step prediction for long-horizon tasks.

\ul{\textbf{Q: How does the action chunk length affect performance?}}

\begin{wrapfigure}{r}{0.48\textwidth}
    \vspace{-0.4em}
    \centering
    \raisebox{0pt}[\dimexpr\height-1.0\baselineskip\relax]{
        \begin{subfigure}[t]{1.0\linewidth}
        \includegraphics[width=0.97\linewidth]{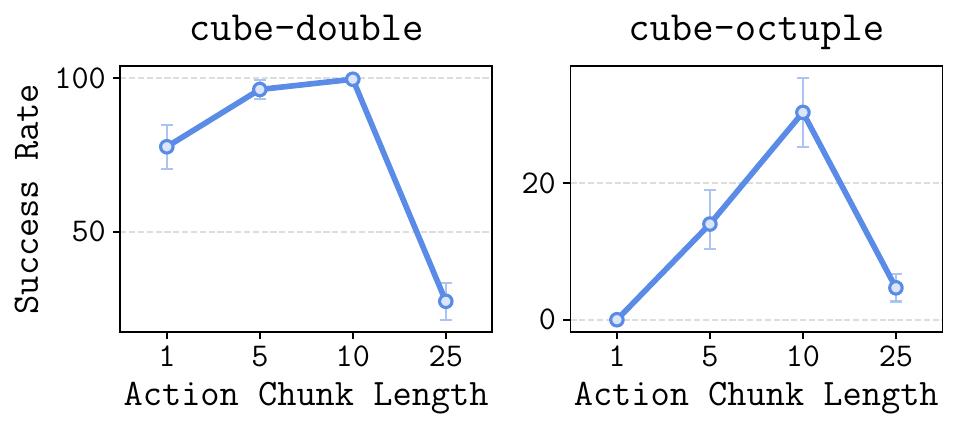}
        \end{subfigure}
    }
    \vspace{-2.0em}
    \caption{
    \footnotesize
    \textbf{Action chunk length vs. performance.}
    }
    \vspace{-1.0em}
    \label{fig:nstep}
\end{wrapfigure}
\textbf{A:}
To answer this question,
we train \methodname with four action chunk lengths ($\{1, 5, 10, 25\}$)
on one short-horizon and one long-horizon task (\tt{cube-double} and \tt{cube-octuple}, respectively)
used in \Cref{sec:exp_long}.
As shown in \Cref{fig:nstep}, action chunking with an appropriate chunk size can substantially improve performance on both tasks.
Notably, while \texttt{cube-double} can still be partially solved without action chunking, \texttt{cube-octuple} cannot be solved at all without it. This demonstrates that action chunking is crucial especially on long-horizon tasks.
However, \Cref{fig:nstep} also shows that too long action chunks can degrade performance,
mainly due to the difficulty of open-loop multi-step future prediction.

\ul{\textbf{Q: How important is flow rejection sampling?}}

\begin{wraptable}{r}{0.48\linewidth}
    \vspace{-1.3em}
    \centering
    \caption{
    \footnotesize
    \textbf{Ablation study of \methodname.}
    }
    \vspace{-1em}
    \vspace{5pt}
    \label{table:abl}
    \raisebox{0pt}[\dimexpr\height-1.0\baselineskip\relax]{
    \scalebox{0.67}{

\begin{tabularew}{l*{3}{>{\spew{.5}{+1}}r@{\,}l}}
\toprule
\multicolumn{1}{l}{\tt{Task}} & \multicolumn{2}{c}{\tt{\methodname (Gau)}} & \multicolumn{2}{c}{\tt{\methodname (FQL)}} & \multicolumn{2}{c}{\tt{\color{myblue}\tt{\methodname}}} \\
\midrule
\tt{cube-single-play-v0} & \tt{2} &{\tiny $\pm$\tt{3}} & \tt{77} &{\tiny $\pm$\tt{21}} & \tt{\color{myblue}{100}} &{\tiny $\pm$\tt{0}} \\
\tt{cube-double-play-v0} & \tt{0} &{\tiny $\pm$\tt{0}} & \tt{2} &{\tiny $\pm$\tt{3}} & \tt{\color{myblue}{50}} &{\tiny $\pm$\tt{12}} \\
\tt{scene-play-v0} & \tt{0} &{\tiny $\pm$\tt{0}} & \tt{40} &{\tiny $\pm$\tt{47}} & \tt{\color{myblue}{100}} &{\tiny $\pm$\tt{0}} \\
\tt{puzzle-3x3-play-v0} & \tt{\color{myblue}{0}} &{\tiny $\pm$\tt{0}} & \tt{\color{myblue}{0}} &{\tiny $\pm$\tt{0}} & \tt{\color{myblue}{0}} &{\tiny $\pm$\tt{0}} \\
\tt{puzzle-4x4-play-v0} & \tt{0} &{\tiny $\pm$\tt{0}} & \tt{23} &{\tiny $\pm$\tt{13}} & \tt{\color{myblue}{85}} &{\tiny $\pm$\tt{14}} \\
\bottomrule
\end{tabularew}

    }
    }
    \vspace{-0.5em}
\end{wraptable}
\textbf{A:}
Another key feature of \methodname is its use of flow rejection sampling.
To understand the importance of this component,
we conduct an ablation study of \methodname by using (1) a Gaussian (``Gau'') action-chunk policy instead of a flow policy,
and (2) gradient-based policy extraction (one-step distillation from FQL~\citep{fql_park2025})
instead of rejection sampling.
We present the ablation results on the default tasks for five reward-based environments used in \Cref{table:abl}.
The results indicate that the use of expressive flow matching is crucial for \methodname,
and that rejection sampling generally yields better performance on most tasks.

\vspace{-0.1em}
\section{Closing Remarks}
\vspace{-0.1em}

In this work, we introduced \methodname
as a model-based actor-critic algorithm that combines an action-chunk policy and an action-chunk model.
\methodname enables generating imaginary autoregressive rollouts up to $100$ steps,
achieving the best performance among model-based RL approaches on challenging, long-horizon tasks.

We now revisit the initial promise of this paper.
In \Cref{sec:intro}, we motivated offline model-based RL as a promising alternative to offline model-free RL
in terms of horizon scalability.
Our answer is (at least partially) affirmative:
on a variety of long-horizon manipulation tasks,
we show that \methodname does outperform state-of-the-art model-free RL algorithms (\Cref{table:long_100m}).
However, as discussed in \Cref{sec:exp_long},
even the best model-based RL algorithm (\methodname)
underperforms on contact-rich locomotion tasks (\eg, \tt{humanoidmaze}),
suggesting room for improvement in sequential dynamics modeling.
We believe that incorporating more advanced modeling techniques could address these limitations,
which we leave for future work.

\section*{Reproducibility Statement}

For the reproducibility of our work, we provide the code of \methodname in \url{https://github.com/kwanyoungpark/MAC}. We fully describe the experimental details and hyperparameters to reproduce the results for our method and baselines in \Cref{sec:exp_details}.

\section*{Acknowledgements}

Kwanyoung Park and Seohong Park are partly supported by the Korea Foundation for Advanced Studies (KFAS). This research used the Savio computational cluster resource provided by the Berkeley Research Computing program at UC Berkeley. This research was partly supported by ONR N00014-22-1-2773 and the National Research Foundation of Korea (NRF) grants funded by the Korean Government (MSIT) (RS-2024-00333634, RS-2025-25396144, RS-2025-25448259).

\bibliography{iclr2026_conference}
\bibliographystyle{iclr2026_conference}

\appendix

\clearpage
\section{Experimental Details}
\label{sec:exp_details}

We implement \methodname on top of the codebase of~\citet{sharsa_park2025}. Each experiment takes approximately $2$ days for large-scale benchmarks, and around $3$ hours for single-task benchmarks on a single A5000 GPU.

\subsection{Implementation Details}

\textbf{Network architectures.} We follow the setup of the work by~\citet{fql_park2025,sharsa_park2025}, using $4$-layer MLPs with layer normalization~\citep{ln_ba2016} for all neural networks (the policy, critic, dynamics model, and reward model). For large-scale benchmarks, we parameterize the reward model and the terminal model using a single success prediction network $f_{\psi}(\pl{s_t}, \pl{\bm{a}_t})$, where termination is calculated as $\mathbbm{1}(f_{\psi}(s_t, \bm{a}_t) > 0.5)$ and reward is calculated as $\mathbbm{1}(f_{\psi}(s_t, \bm{a}_t) > 0.5) - 1$. For reward-based tasks, we use a reward model $r_{\psi}(\pl{s_t}, \pl{\bm{a}_t})$ without termination.

\textbf{Accelerating rejection sampling during training.}
% A key drawback of rejection sampling is its computational cost; if we sample $N$ candidates, generating a rollout of length $H$ requires $HN$ critic evaluations.
To improve training time, we use different numbers of samples for rejection sampling during training and evaluation (which we call $N_\mathrm{train}$ and $N_\mathrm{test}$.
% Interestingly, we found it sufficient to use a smaller number of samples during training,
Specifically,
we use $N_{\mathrm{train}}=8$ during training (except in \texttt{puzzle-4x5}, where a larger $N_{\mathrm{train}}=32$ was necessary due to the BC policy branching over $20$ possible actions)
% While a smaller $N$ is effective for training,
and $N_{\mathrm{test}}=32$ at test time. %is crucial for reliable performance across most tasks. %TODO

\textbf{Implementation details for the compared methods.} 
We implement MOPO, MOBILE, and LEQ in our codebase. For MOPO, epistemic uncertainty is estimated as the maximum standard deviation across ensemble members~\citep{mopo_yu2020}. For LEQ, we omit dataset expansion, which we found to have a negligible impact in our benchmarks. We use $5$ dynamics model ensembles for all methods and disable early stopping and validation filtering when training the model, as we found they are unreliable on large-scale datasets (training and validation metrics are nearly identical in these settings).

For D-MPC~\citep{dmpc_zhou2025}, we implement the flow variant of D-MPC (F-MPC) in our codebase. Specifically, we train a flow BC policy $\pi (\pl{a_{t:t+n-1}} \mid \pl{s_t})$ and a flow dynamics model $p_{\psi}(\pl{s_{t+1:t+n}} \mid \pl{s_t, a_{t:t+n-1}})$ instead of using diffusion models. For reward-based benchmarks, we calculate the return-to-go as $G_t = \sum_{t'=t}^{T} r_{t'}$ without discounts, as in the original paper. For goal-conditioned (large-scale) benchmarks, we similarly define the return-to-go without discounts for the goal-conditioned tasks as $G_t = \mathbbm{1}(g \in \{s_t, \cdots, s_T\})$. 
Unlike the original architecture, we do not use history conditioning and transformers (as all tasks are Markovian) and use the same MLP architecture as other methods for a fair comparison.

For sequence modeling approaches (Diffuser and HD-DA), we follow the official implementation for D4RL's \tt{maze2d} environment~\citep{d4rl_fu2020}, and adjust the maximum length of the trajectory generation and the number of diffusion steps (of the high-level policy for HD-DA) to be the maximum length of the environment (\eg, $H=4000$ for \tt{humanoidmaze-giant}). We re-plan the trajectory every $100$ steps, as we found that this is necessary to achieve a non-zero performance on long-horizon tasks, unlike in the \texttt{maze2d} benchmark.

For other model-free methods, we use the implementations by \citet{sharsa_park2025} and~\citet{ogbench_park2025}. We also take the results from these papers for the corresponding methods.

\textbf{Implementation details for ablation experiments.}
For the ablation study on the action-chunk length, we fix the horizon length $H$ to $10$ and only change the action-chunk length $n \in \{1, 5, 10, 25\}$. For MAC (Gau) of the ablation study on flow rejection sampling, we parameterize the action-chunk policy with $\bm{a}_t = \tanh(x_t)$, where $x_t \sim \mathcal{N}(\mu_\theta (s_t), \sigma^2_{\theta}(s_t))$.

\begin{figure*}[b!]
    \vspace{-1.5em}
    \centering
    \includegraphics[width=\textwidth]{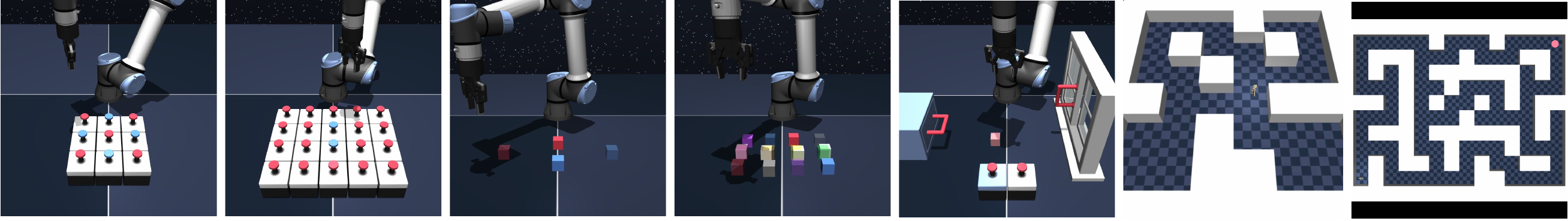}
    \vspace{-1.5em}
    \caption{\textbf{OGBench tasks.}
    %\sh{Try ``Align $\to$ Distribute horizontally`` in PowerPoint to ensure equal spacing between figures. Also, is there a reason to put \tt{scene} in the rightmost column (given its higher similarity to other manipulation environments on the left)?}
    }
    %\vspace{-2em}
\end{figure*}

\clearpage
\subsection{Training Time}
\label{sec:time}
We report the average training time and inference time for single-task and multi-task experiments in A5000 for MAC and prior MBRL methods in the table below. MAC trains in around 3 hours for a single task and 55 hours for multi-task experiments, which is 1.2 - 2.2 times longer than other methods. Inference speed of MAC is similar or 1.5 times longer than other methods. All models use identical architecture sizes across methods.

\begin{table}[H]
    \small
    \centering
    \caption{\footnotesize\textbf{Training time of \methodname and prior MBRL methods.}}
    \label{tab:training_time}
    \begin{tabular}{lccccc}
    \toprule
    \textbf{Training time (hours)} & \texttt{MOPO} & \texttt{MOBILE} & \texttt{FMPC} & \texttt{LEQ (H=5)} & \texttt{MAC} \\
    \midrule
    \textbf{Single-task} & 1.4 & 2.6 & 1.7 & 1.6 & 3.1  \\
    \textbf{Multi-task} & 25.1 & 36.7 & 28.4 & 25.2 & 55.5  \\
    \bottomrule
    \end{tabular}
\end{table}

\begin{table}[H]
    \small
    \centering
    \caption{\footnotesize\textbf{Inference time of \methodname and prior MBRL methods.}}
    \label{tab:training_time}
    \begin{tabular}{lccccc}
    \toprule
    \textbf{Inference time (ms)} & \texttt{MOPO} & \texttt{MOBILE} & \texttt{FMPC} & \texttt{LEQ} & \texttt{MAC} \\
    \midrule
    \textbf{Single-task} & 1.8 & 1.8 & 2.3 & 1.6 & 2.5  \\
    \textbf{Multi-task} & 5.1 & 5.2 & 7.3 & 5.2 & 7.2  \\
    \bottomrule
    \end{tabular}
\end{table}

\subsection{Hyperparameters}

\textbf{Shared hyperparameters.}
Here, we report shared hyperparameters for \methodname, MOPO, MOBILE, and all model-free baselines.
% We report the hyperparameters shared for all methods we implemented in, or compatible with our codebase (namely, \methodname, MOPO, MOBILE, and all model-free baselines).
The hyperparameters for goal-conditioned tasks are presented in \Cref{tab:hyperparameters_multitask}, and those for reward-based tasks are in \Cref{tab:hyperparameters_singletask}. We note that these hyperparameter configurations mostly follow those of SHARSA~\citep{sharsa_park2025} for multi-task experiments, and FQL~\citep{fql_park2025} for single-task experiments.

\begin{table}[H]
    \small
    \centering
    \caption{\footnotesize\textbf{Shared hyperparameters for large-scale benchmark tasks.}}
    \label{tab:hyperparameters_multitask}
    \begin{tabular}{ll}
    \toprule
    \textbf{Hyperparameters} & \textbf{Value} \\
    \midrule
    Gradient steps  & $2$M  \\
    Learning rate   & $3 \times 10^{-4}$  \\
    Optimizer       & Adam~\citep{adam_kingma2015}  \\
    Batch size & $1024$  \\
    MLP size & $[1024, 1024, 1024, 1024]$ \\
    Actor $(p^\gD_\mathrm{cur}, p^\gD_\mathrm{geom}, p^\gD_\mathrm{traj}, p^\gD_\mathrm{rand})$ ratio 
    & \makecell[lt]{$(0, 1, 0, 0)$ (\tt{cube}) \\ $(0, 0.5, 0, 0.5)$ (\tt{puzzle}) \\ $(0, 0, 1, 0)$ (\tt{humanoidmaze}) }\\
    Value $(p^\gD_\mathrm{cur}, p^\gD_\mathrm{geom}, p^\gD_\mathrm{traj}, p^\gD_\mathrm{rand})$ ratio 
    & $(0.2, 0, 0.5, 0.3)$ \\
    \bottomrule
    \end{tabular}
\end{table}

\begin{table}[H]
    \small
    \centering
    \caption{\footnotesize\textbf{Shared hyperparameters for reward-based benchmark tasks.}}
    \label{tab:hyperparameters_singletask}
    \begin{tabular}{ll}
    \toprule
    \textbf{Hyperparameters} & \textbf{Value} \\
    \midrule
    Gradient steps  & $1$M  \\
    Learning rate   & $3 \times 10^{-4}$  \\
    Optimizer       & Adam~\citep{adam_kingma2015}  \\
    Batch size & $256$  \\
    MLP size & $[512, 512, 512, 512]$ \\
    \bottomrule
    \end{tabular}
\end{table}

\paragraph{\methodname hyperparameters.}
We report the hyperparameters for our method in \Cref{tab:hyperparameters}.
Note that \methodname uses the same $(\achunk, H, N_\mathrm{train}, N_\mathrm{test}) = (10, 10, 8, 32)$ across all tasks,
except for \texttt{puzzle-4x5}, where using $N_{\mathrm{train}}=32$ during training is important
as the BC policy has $20$ possible branches.
Accordingly, only in \tt{puzzle-4x5},
we decrease the hidden dimensionality of the networks to $256$
to compensate for the increased training time from $N_\mathrm{train}=32$.
% because the original network was too slow with $N_{\mathrm{train}}=32$. 

\begin{table}[H]
    \small
    \centering
    \caption{\footnotesize\textbf{Hyperparameters of \methodname.}}
    \label{tab:hyperparameters}
    \begin{tabular}{ll}
    \toprule
    \textbf{Hyperparameters} & \textbf{Value} \\
    \midrule
    Learning rate   & $3 \times 10^{-4}$  \\
    Optimizer       & Adam  \\
    Nonlinearity & GELU~\citep{gelu_hendrycks2016} \\
    Layer normalization & True \\
    Target network update rate & $0.005$ \\
    Discount factor $\gamma$ & $0.999$ \\
    Flow steps & $10$ \\
    $N_{\mathrm{train}}$  & $8$ (default), $32$ (\texttt{puzzle-4x5}) \\
    $N_{\mathrm{test}}$  & $32$ \\
    Rollout length $H$  & $10$   \\
    Action-chunk size $\achunk$ & $10$  \\
    \bottomrule
    \end{tabular}
\end{table}

\textbf{Hyperparameters for baselines.}
We report the optimal hyperparameters of all baselines for goal-conditioned experiments in \Cref{tab:baseline_hyperparameters_multitask} and reward-based experiments in \Cref{tab:baseline_hyperparameters_singletask}.

For MOPO and MOBILE, we perform a hyperparameter sweep over rollout lengths $H \in \{1, 5, 10\}$ and penalty coefficients $\beta \in \{0.1, 0.5, 1.0, 2.0, 3.0, 5.0\}$, where $H$ denotes the model rollout horizon and $\beta$ is the penalization coefficient for model uncertainty or Bellman inconsistency, respectively. We note that reducing the MBPO loop’s model batch ratio $f$ from $0.95$ to $f \in \{0.5, 0.25\}$ is crucial for training on long-horizon tasks, as also noted by~\citet{leq_park2025}. 

For LEQ, we search over rollout lengths $H \in \{1, 5, 10\}$ and expectiles $\tau \in \{0.1, 0.3, 0.5\}$, where the expectile $\tau$ controls the degree of conservatism for critic and policy learning.

For model-free methods in large-scale benchmarks, we follow the list of hyperparameters to search over in the work by \citet{sharsa_park2025}. For SHARSA, we searched over $n \in \{25, 50\}$. For n-SAC+BC, we search over $n \in {10, 25, 50}$ and regularization coefficients $\alpha \in \{0.01, 0.03, 0.1, 0.3\}$. For GCIQL, we follow \citep{bottleneck_park2024} and extract policies with DDPG+BC, searching over $\alpha \in \{0.003, 0.01, 0.03, 0.1, 0.3, 1.0, 3.0\}$.

We denote ``N/A'' in the tables if a method achieves zero performance across all hyperparameters tested in our sweep. If not specified, all other hyperparameters follow the defaults provided in the original papers.

\begin{table}[h]
    \small
    \centering
    \caption{\footnotesize\textbf{Hyperparameters for baselines for large-scale benchmark tasks.}}
    \label{tab:baseline_hyperparameters_multitask}
    \begin{tabular}{llll}
    \toprule
    \tt{Environment} & 
    \tt{MOPO} $(H, \beta, f)$ & \tt{MOBILE} $(H, \beta, f)$ & \tt{LEQ} $(H, \tau)$ \\
    \midrule
    \texttt{cube-double-play-v0}& $(10, 1.0, 0.25)$ & $(5, 0.5, 0.5)$ & N/A \\
    \texttt{cube-octuple-play-v0}& N/A & N/A  & N/A \\
    \texttt{humanoidmaze-medium-navigate-v0} & $(1, 0.5, 0.5)$ & $(1, 1.0, 0.5)$ & N/A \\
    \texttt{humanoidmaze-giant-navigate-v0}      & N/A  & N/A  & N/A \\
    \texttt{puzzle-3x3-play-v0} & $(5, 5.0, 0.5)$ & $(10, 3.0, 0.25)$ & $(1, 0.1)$ \\
    \texttt{puzzle-4x5-play-v0} & N/A  & N/A  & $(1, 0.1)$ \\
    \bottomrule
    \end{tabular}
    %\vspace{-0.75em}
\end{table}

\begin{table}[h]
    \small
    \centering
    \caption{\footnotesize\textbf{Hyperparameters for baselines for reward-based benchmark tasks.}}
    \label{tab:baseline_hyperparameters_singletask}
    \begin{tabular}{llll}
    \toprule
    \tt{Environment} & 
    \tt{MOPO} $(H, \beta, f)$ & \tt{MOBILE} $(H, \beta, f)$ & \tt{LEQ} $(H, \tau)$ \\
    \midrule
    %Search range & \makecell[l]{$H=\{1, 5, 10\}$ \\ $\beta = \{0.5, 1.0, 2.0, 5.0\}$} &  \makecell[l]{$H=\{1, 5, 10\}$ \\ $\beta = \{0.5, 1.0, 2.0, 5.0\}$} & \makecell[l]{$H=\{1, 5, 10\}$ \\ $\tau = \{0.1, 0.3, 0.5\}$} \\
    \texttt{cube-single-play-v0}& $(10, 2.0, 0.25)$ & $(10, 5.0, 0.25)$ & N/A \\
    \texttt{cube-double-play-v0}& $(10, 1.0, 0.25)$ & N/A  & N/A \\
    \texttt{scene-play-v0}      & $(10, 2.0, 0.25)$  & N/A  & N/A \\
    \texttt{puzzle-3x3-play-v0} & N/A  & N/A  & N/A \\
    \texttt{puzzle-4x4-play-v0} & N/A  & N/A  & N/A \\
    \bottomrule
    \end{tabular}
    %\vspace{-0.75em}
\end{table}

\textbf{Hyperparameters for ablation studies.}
We report the optimal hyperparameters for the ablated variants of \methodname:
one that replaces the flow policy with a Gaussian policy (``Gau''),
and another that replaces rejection sampling with FQL's one-step distillation (``FQL'').
For the Gaussian policy variant, we reuse the same hyperparameters as our main method.
For the FQL variant, we search over the behavior cloning coefficients $\alpha \in \{0.1, 0.3, 1.0, 3.0\}$.

\begin{table}[h]
    \small
    \centering
    \caption{\footnotesize\textbf{Hyperparameters for ablation experiments.}}
    \label{tab:baseline_hyperparameters_singletask}
    \begin{tabular}{ll}
    \toprule
    \tt{Environment} & 
    \tt{\methodname (FQL)} ($\alpha$) \\
    \midrule
    %Search range & \makecell[l]{$H=\{1, 5, 10\}$ \\ $\beta = \{0.5, 1.0, 2.0, 5.0\}$} &  \makecell[l]{$H=\{1, 5, 10\}$ \\ $\beta = \{0.5, 1.0, 2.0, 5.0\}$} & \makecell[l]{$H=\{1, 5, 10\}$ \\ $\tau = \{0.1, 0.3, 0.5\}$} \\
    \texttt{cube-single-play-v0}& $1.0$ \\
    \texttt{cube-double-play-v0}& $0.3$ \\
    \texttt{scene-play-v0}      & $1.0$ \\
    \texttt{puzzle-3x3-play-v0} & $1.0$ \\
    \texttt{puzzle-4x4-play-v0} & $1.0$ \\
    \bottomrule
    \end{tabular}
    %\vspace{-0.75em}
\end{table}

\clearpage
\section{Complete Numerical Results}
\label{sec:number_results}
For completeness, we provide the full per-task results for large-scale, long-horizon environments and reward-based environments in \Cref{table:full_long_100m} and \Cref{table:full_standard} (corresponding to \Cref{table:long_100m} and \Cref{table:standard}). The results are averaged over $4$ seeds and we report the standard deviations for each tasks. We highlight the numbers that are above or equal to $95\%$ of the best performance.
%Specifically, we report each method's average success rate ($\%$) across the five goals on each task.
% The results are averaged over $4$ seeds and we report the standard deviations after the $\pm$ sign. Numbers with the best values are bolded.

\begin{table}[h!]
\caption{
\footnotesize \textbf{Complete results for large-scale experiments.}
}
\label{table:full_long_100m}

\centering
\scalebox{0.62}
{
\begin{tabularew}{ll*{10}{>{\spew{.5}{+1}}r@{\,}l}}
\toprule
& \multicolumn{1}{c}{} & \multicolumn{6}{c}{\tt{Model-Free}} & \multicolumn{4}{c}{\tt{Seq. Modeling}} & \multicolumn{10}{c}{\tt{Model-Based}} \\
\cmidrule(r){3-8}
\cmidrule(r){9-12}
\cmidrule(r){13-22}

\multicolumn{1}{l}{\tt{Environment}} & \multicolumn{1}{l}{\tt{Task}} & \multicolumn{2}{c}{\tt{GCIQL}} & \multicolumn{2}{c}{\tt{n-SAC+BC}} & \multicolumn{2}{c}{\tt{SHARSA}} & \multicolumn{2}{c}{\tt{Diffuser}} & \multicolumn{2}{c}{\tt{HD-DA}} & \multicolumn{2}{c}{\tt{MOPO}} & \multicolumn{2}{c}{\tt{MOBILE}} & \multicolumn{2}{c}{\tt{LEQ}} & \multicolumn{2}{c}{\tt{FMPC}} & \multicolumn{2}{c}{\tt{\color{myblue}\tt{MAC}}} \\
\midrule
\multirow[c]{6}{*}{\makecell[l]{\tt{humanoidmaze-medium-navigate-} \\ \tt{oraclerep-v0}}}&\tt{task1}& \tt{82} &{\tiny $\pm$\tt{3}} & \tt{\color{myblue}{97}} &{\tiny $\pm$\tt{4}} & \tt{\color{myblue}{95}} &{\tiny $\pm$\tt{6}} & \tt{0} &{\tiny $\pm$\tt{0}} & \tt{0} &{\tiny $\pm$\tt{0}} & \tt{48} &{\tiny $\pm$\tt{27}} & \tt{38} &{\tiny $\pm$\tt{14}} & \tt{0} &{\tiny $\pm$\tt{0}} & \tt{27} &{\tiny $\pm$\tt{9}} & \tt{67} &{\tiny $\pm$\tt{12}} \\
 &\tt{task2}& \tt{\color{myblue}{95}} &{\tiny $\pm$\tt{6}} & \tt{\color{myblue}{100}} &{\tiny $\pm$\tt{0}} & \tt{\color{myblue}{100}} &{\tiny $\pm$\tt{0}} & \tt{0} &{\tiny $\pm$\tt{0}} & \tt{0} &{\tiny $\pm$\tt{0}} & \tt{85} &{\tiny $\pm$\tt{21}} & \tt{75} &{\tiny $\pm$\tt{15}} & \tt{0} &{\tiny $\pm$\tt{0}} & \tt{22} &{\tiny $\pm$\tt{11}} & \tt{87} &{\tiny $\pm$\tt{9}} \\
 &\tt{task3}& \tt{0} &{\tiny $\pm$\tt{0}} & \tt{\color{myblue}{98}} &{\tiny $\pm$\tt{3}} & \tt{\color{myblue}{100}} &{\tiny $\pm$\tt{0}} & \tt{0} &{\tiny $\pm$\tt{0}} & \tt{0} &{\tiny $\pm$\tt{0}} & \tt{0} &{\tiny $\pm$\tt{0}} & \tt{0} &{\tiny $\pm$\tt{0}} & \tt{0} &{\tiny $\pm$\tt{0}} & \tt{18} &{\tiny $\pm$\tt{3}} & \tt{7} &{\tiny $\pm$\tt{0}} \\
 &\tt{task4}& \tt{0} &{\tiny $\pm$\tt{0}} & \tt{\color{myblue}{97}} &{\tiny $\pm$\tt{4}} & \tt{82} &{\tiny $\pm$\tt{3}} & \tt{0} &{\tiny $\pm$\tt{0}} & \tt{0} &{\tiny $\pm$\tt{0}} & \tt{0} &{\tiny $\pm$\tt{0}} & \tt{0} &{\tiny $\pm$\tt{0}} & \tt{0} &{\tiny $\pm$\tt{0}} & \tt{5} &{\tiny $\pm$\tt{6}} & \tt{0} &{\tiny $\pm$\tt{0}} \\
 &\tt{task5}& \tt{\color{myblue}{98}} &{\tiny $\pm$\tt{3}} & \tt{\color{myblue}{98}} &{\tiny $\pm$\tt{3}} & \tt{\color{myblue}{100}} &{\tiny $\pm$\tt{0}} & \tt{0} &{\tiny $\pm$\tt{0}} & \tt{0} &{\tiny $\pm$\tt{0}} & \tt{0} &{\tiny $\pm$\tt{0}} & \tt{0} &{\tiny $\pm$\tt{0}} & \tt{0} &{\tiny $\pm$\tt{0}} & \tt{20} &{\tiny $\pm$\tt{11}} & \tt{22} &{\tiny $\pm$\tt{14}} \\
 &\tt{overall}& \tt{55} &{\tiny $\pm$\tt{1}} & \tt{\color{myblue}{98}} &{\tiny $\pm$\tt{2}} & \tt{\color{myblue}{95}} &{\tiny $\pm$\tt{2}} & \tt{0} &{\tiny $\pm$\tt{0}} & \tt{0} &{\tiny $\pm$\tt{0}} & \tt{27} &{\tiny $\pm$\tt{5}} & \tt{23} &{\tiny $\pm$\tt{3}} & \tt{0} &{\tiny $\pm$\tt{0}} & \tt{18} &{\tiny $\pm$\tt{5}} & \tt{36} &{\tiny $\pm$\tt{2}} \\
\midrule
\multirow[c]{6}{*}{\makecell[l]{\tt{humanoidmaze-giant-navigate-} \\ \tt{oraclerep-v0}}}&\tt{task1}& \tt{0} &{\tiny $\pm$\tt{0}} & \tt{\color{myblue}{58}} &{\tiny $\pm$\tt{18}} & \tt{22} &{\tiny $\pm$\tt{18}} & \tt{0} &{\tiny $\pm$\tt{0}} & \tt{0} &{\tiny $\pm$\tt{0}} & \tt{0} &{\tiny $\pm$\tt{0}} & \tt{0} &{\tiny $\pm$\tt{0}} & \tt{0} &{\tiny $\pm$\tt{0}} & \tt{0} &{\tiny $\pm$\tt{0}} & \tt{0} &{\tiny $\pm$\tt{0}} \\
 &\tt{task2}& \tt{10} &{\tiny $\pm$\tt{7}} & \tt{\color{myblue}{87}} &{\tiny $\pm$\tt{8}} & \tt{43} &{\tiny $\pm$\tt{22}} & \tt{0} &{\tiny $\pm$\tt{0}} & \tt{0} &{\tiny $\pm$\tt{0}} & \tt{0} &{\tiny $\pm$\tt{0}} & \tt{0} &{\tiny $\pm$\tt{0}} & \tt{0} &{\tiny $\pm$\tt{0}} & \tt{0} &{\tiny $\pm$\tt{0}} & \tt{0} &{\tiny $\pm$\tt{0}} \\
 &\tt{task3}& \tt{5} &{\tiny $\pm$\tt{3}} & \tt{\color{myblue}{85}} &{\tiny $\pm$\tt{11}} & \tt{23} &{\tiny $\pm$\tt{19}} & \tt{0} &{\tiny $\pm$\tt{0}} & \tt{0} &{\tiny $\pm$\tt{0}} & \tt{0} &{\tiny $\pm$\tt{0}} & \tt{0} &{\tiny $\pm$\tt{0}} & \tt{0} &{\tiny $\pm$\tt{0}} & \tt{0} &{\tiny $\pm$\tt{0}} & \tt{0} &{\tiny $\pm$\tt{0}} \\
 &\tt{task4}& \tt{2} &{\tiny $\pm$\tt{3}} & \tt{\color{myblue}{82}} &{\tiny $\pm$\tt{11}} & \tt{40} &{\tiny $\pm$\tt{14}} & \tt{0} &{\tiny $\pm$\tt{0}} & \tt{0} &{\tiny $\pm$\tt{0}} & \tt{0} &{\tiny $\pm$\tt{0}} & \tt{0} &{\tiny $\pm$\tt{0}} & \tt{0} &{\tiny $\pm$\tt{0}} & \tt{0} &{\tiny $\pm$\tt{0}} & \tt{0} &{\tiny $\pm$\tt{0}} \\
 &\tt{task5}& \tt{3} &{\tiny $\pm$\tt{4}} & \tt{\color{myblue}{98}} &{\tiny $\pm$\tt{3}} & \tt{87} &{\tiny $\pm$\tt{18}} & \tt{0} &{\tiny $\pm$\tt{0}} & \tt{0} &{\tiny $\pm$\tt{0}} & \tt{0} &{\tiny $\pm$\tt{0}} & \tt{0} &{\tiny $\pm$\tt{0}} & \tt{0} &{\tiny $\pm$\tt{0}} & \tt{0} &{\tiny $\pm$\tt{0}} & \tt{0} &{\tiny $\pm$\tt{0}} \\
 &\tt{overall}& \tt{4} &{\tiny $\pm$\tt{2}} & \tt{\color{myblue}{82}} &{\tiny $\pm$\tt{5}} & \tt{43} &{\tiny $\pm$\tt{6}} & \tt{0} &{\tiny $\pm$\tt{0}} & \tt{0} &{\tiny $\pm$\tt{0}} & \tt{0} &{\tiny $\pm$\tt{0}} & \tt{0} &{\tiny $\pm$\tt{0}} & \tt{0} &{\tiny $\pm$\tt{0}} & \tt{0} &{\tiny $\pm$\tt{0}} & \tt{0} &{\tiny $\pm$\tt{0}} \\
\midrule
\multirow[c]{6}{*}{\tt{cube-double-play-oraclerep-v0} }&\tt{task1}& \tt{\color{myblue}{100}} &{\tiny $\pm$\tt{0}} & \tt{67} &{\tiny $\pm$\tt{32}} & \tt{\color{myblue}{100}} &{\tiny $\pm$\tt{0}} & \tt{6} &{\tiny $\pm$\tt{3}} & \tt{6} &{\tiny $\pm$\tt{3}} & \tt{42} &{\tiny $\pm$\tt{18}} & \tt{50} &{\tiny $\pm$\tt{12}} & \tt{0} &{\tiny $\pm$\tt{0}} & \tt{73} &{\tiny $\pm$\tt{14}} & \tt{\color{myblue}{100}} &{\tiny $\pm$\tt{0}} \\
 &\tt{task2}& \tt{\color{myblue}{100}} &{\tiny $\pm$\tt{0}} & \tt{13} &{\tiny $\pm$\tt{11}} & \tt{\color{myblue}{100}} &{\tiny $\pm$\tt{0}} & \tt{0} &{\tiny $\pm$\tt{0}} & \tt{0} &{\tiny $\pm$\tt{1}} & \tt{17} &{\tiny $\pm$\tt{14}} & \tt{15} &{\tiny $\pm$\tt{11}} & \tt{0} &{\tiny $\pm$\tt{0}} & \tt{37} &{\tiny $\pm$\tt{28}} & \tt{\color{myblue}{100}} &{\tiny $\pm$\tt{0}} \\
 &\tt{task3}& \tt{\color{myblue}{100}} &{\tiny $\pm$\tt{0}} & \tt{37} &{\tiny $\pm$\tt{23}} & \tt{\color{myblue}{100}} &{\tiny $\pm$\tt{0}} & \tt{0} &{\tiny $\pm$\tt{0}} & \tt{0} &{\tiny $\pm$\tt{1}} & \tt{18} &{\tiny $\pm$\tt{15}} & \tt{7} &{\tiny $\pm$\tt{5}} & \tt{0} &{\tiny $\pm$\tt{0}} & \tt{43} &{\tiny $\pm$\tt{23}} & \tt{\color{myblue}{100}} &{\tiny $\pm$\tt{0}} \\
 &\tt{task4}& \tt{33} &{\tiny $\pm$\tt{14}} & \tt{15} &{\tiny $\pm$\tt{18}} & \tt{73} &{\tiny $\pm$\tt{14}} & \tt{0} &{\tiny $\pm$\tt{0}} & \tt{0} &{\tiny $\pm$\tt{1}} & \tt{20} &{\tiny $\pm$\tt{14}} & \tt{0} &{\tiny $\pm$\tt{0}} & \tt{0} &{\tiny $\pm$\tt{0}} & \tt{3} &{\tiny $\pm$\tt{4}} & \tt{\color{myblue}{98}} &{\tiny $\pm$\tt{3}} \\
 &\tt{task5}& \tt{38} &{\tiny $\pm$\tt{16}} & \tt{28} &{\tiny $\pm$\tt{33}} & \tt{\color{myblue}{100}} &{\tiny $\pm$\tt{0}} & \tt{0} &{\tiny $\pm$\tt{1}} & \tt{1} &{\tiny $\pm$\tt{0}} & \tt{30} &{\tiny $\pm$\tt{13}} & \tt{5} &{\tiny $\pm$\tt{3}} & \tt{0} &{\tiny $\pm$\tt{0}} & \tt{30} &{\tiny $\pm$\tt{16}} & \tt{\color{myblue}{100}} &{\tiny $\pm$\tt{0}} \\
 &\tt{overall}& \tt{74} &{\tiny $\pm$\tt{3}} & \tt{32} &{\tiny $\pm$\tt{20}} & \tt{\color{myblue}{95}} &{\tiny $\pm$\tt{3}} & \tt{1} &{\tiny $\pm$\tt{1}} & \tt{2} &{\tiny $\pm$\tt{1}} & \tt{25} &{\tiny $\pm$\tt{12}} & \tt{15} &{\tiny $\pm$\tt{3}} & \tt{0} &{\tiny $\pm$\tt{0}} & \tt{37} &{\tiny $\pm$\tt{13}} & \tt{\color{myblue}{100}} &{\tiny $\pm$\tt{1}} \\
\midrule
\multirow[c]{6}{*}{\tt{cube-octuple-play-oraclerep-v0} }&\tt{task1}& \tt{0} &{\tiny $\pm$\tt{0}} & \tt{0} &{\tiny $\pm$\tt{0}} & \tt{\color{myblue}{88}} &{\tiny $\pm$\tt{6}} & \tt{0} &{\tiny $\pm$\tt{0}} & \tt{0} &{\tiny $\pm$\tt{0}} & \tt{0} &{\tiny $\pm$\tt{0}} & \tt{0} &{\tiny $\pm$\tt{0}} & \tt{0} &{\tiny $\pm$\tt{0}} & \tt{0} &{\tiny $\pm$\tt{0}} & \tt{83} &{\tiny $\pm$\tt{4}} \\
 &\tt{task2}& \tt{0} &{\tiny $\pm$\tt{0}} & \tt{0} &{\tiny $\pm$\tt{0}} & \tt{5} &{\tiny $\pm$\tt{10}} & \tt{0} &{\tiny $\pm$\tt{0}} & \tt{0} &{\tiny $\pm$\tt{0}} & \tt{0} &{\tiny $\pm$\tt{0}} & \tt{0} &{\tiny $\pm$\tt{0}} & \tt{0} &{\tiny $\pm$\tt{0}} & \tt{0} &{\tiny $\pm$\tt{0}} & \tt{\color{myblue}{20}} &{\tiny $\pm$\tt{9}} \\
 &\tt{task3}& \tt{0} &{\tiny $\pm$\tt{0}} & \tt{0} &{\tiny $\pm$\tt{0}} & \tt{3} &{\tiny $\pm$\tt{7}} & \tt{0} &{\tiny $\pm$\tt{0}} & \tt{0} &{\tiny $\pm$\tt{0}} & \tt{0} &{\tiny $\pm$\tt{0}} & \tt{0} &{\tiny $\pm$\tt{0}} & \tt{0} &{\tiny $\pm$\tt{0}} & \tt{0} &{\tiny $\pm$\tt{0}} & \tt{\color{myblue}{40}} &{\tiny $\pm$\tt{21}} \\
 &\tt{task4}& \tt{0} &{\tiny $\pm$\tt{0}} & \tt{0} &{\tiny $\pm$\tt{0}} & \tt{0} &{\tiny $\pm$\tt{0}} & \tt{0} &{\tiny $\pm$\tt{0}} & \tt{0} &{\tiny $\pm$\tt{0}} & \tt{0} &{\tiny $\pm$\tt{0}} & \tt{0} &{\tiny $\pm$\tt{0}} & \tt{0} &{\tiny $\pm$\tt{0}} & \tt{0} &{\tiny $\pm$\tt{0}} & \tt{\color{myblue}{5}} &{\tiny $\pm$\tt{6}} \\
 &\tt{task5}& \tt{0} &{\tiny $\pm$\tt{0}} & \tt{0} &{\tiny $\pm$\tt{0}} & \tt{0} &{\tiny $\pm$\tt{0}} & \tt{0} &{\tiny $\pm$\tt{0}} & \tt{0} &{\tiny $\pm$\tt{0}} & \tt{0} &{\tiny $\pm$\tt{0}} & \tt{0} &{\tiny $\pm$\tt{0}} & \tt{0} &{\tiny $\pm$\tt{0}} & \tt{0} &{\tiny $\pm$\tt{0}} & \tt{\color{myblue}{3}} &{\tiny $\pm$\tt{4}} \\
 &\tt{overall}& \tt{0} &{\tiny $\pm$\tt{0}} & \tt{0} &{\tiny $\pm$\tt{0}} & \tt{19} &{\tiny $\pm$\tt{3}} & \tt{0} &{\tiny $\pm$\tt{0}} & \tt{0} &{\tiny $\pm$\tt{0}} & \tt{0} &{\tiny $\pm$\tt{0}} & \tt{0} &{\tiny $\pm$\tt{0}} & \tt{0} &{\tiny $\pm$\tt{0}} & \tt{0} &{\tiny $\pm$\tt{0}} & \tt{\color{myblue}{30}} &{\tiny $\pm$\tt{6}} \\
\midrule
\multirow[c]{6}{*}{\tt{puzzle-3x3-play-oraclerep-v0} }&\tt{task1}& \tt{\color{myblue}{100}} &{\tiny $\pm$\tt{0}} & \tt{\color{myblue}{95}} &{\tiny $\pm$\tt{6}} & \tt{\color{myblue}{100}} &{\tiny $\pm$\tt{0}} & \tt{3} &{\tiny $\pm$\tt{2}} & \tt{4} &{\tiny $\pm$\tt{4}} & \tt{93} &{\tiny $\pm$\tt{9}} & \tt{77} &{\tiny $\pm$\tt{23}} & \tt{3} &{\tiny $\pm$\tt{7}} & \tt{25} &{\tiny $\pm$\tt{15}} & \tt{\color{myblue}{100}} &{\tiny $\pm$\tt{0}} \\
 &\tt{task2}& \tt{\color{myblue}{100}} &{\tiny $\pm$\tt{0}} & \tt{80} &{\tiny $\pm$\tt{11}} & \tt{\color{myblue}{100}} &{\tiny $\pm$\tt{0}} & \tt{0} &{\tiny $\pm$\tt{0}} & \tt{0} &{\tiny $\pm$\tt{1}} & \tt{0} &{\tiny $\pm$\tt{0}} & \tt{0} &{\tiny $\pm$\tt{0}} & \tt{0} &{\tiny $\pm$\tt{0}} & \tt{18} &{\tiny $\pm$\tt{18}} & \tt{\color{myblue}{100}} &{\tiny $\pm$\tt{0}} \\
 &\tt{task3}& \tt{\color{myblue}{98}} &{\tiny $\pm$\tt{3}} & \tt{93} &{\tiny $\pm$\tt{5}} & \tt{\color{myblue}{100}} &{\tiny $\pm$\tt{0}} & \tt{0} &{\tiny $\pm$\tt{0}} & \tt{0} &{\tiny $\pm$\tt{0}} & \tt{0} &{\tiny $\pm$\tt{0}} & \tt{0} &{\tiny $\pm$\tt{0}} & \tt{0} &{\tiny $\pm$\tt{0}} & \tt{8} &{\tiny $\pm$\tt{8}} & \tt{\color{myblue}{100}} &{\tiny $\pm$\tt{0}} \\
 &\tt{task4}& \tt{\color{myblue}{100}} &{\tiny $\pm$\tt{0}} & \tt{92} &{\tiny $\pm$\tt{8}} & \tt{\color{myblue}{100}} &{\tiny $\pm$\tt{0}} & \tt{0} &{\tiny $\pm$\tt{0}} & \tt{0} &{\tiny $\pm$\tt{0}} & \tt{0} &{\tiny $\pm$\tt{0}} & \tt{0} &{\tiny $\pm$\tt{0}} & \tt{0} &{\tiny $\pm$\tt{0}} & \tt{2} &{\tiny $\pm$\tt{3}} & \tt{\color{myblue}{100}} &{\tiny $\pm$\tt{0}} \\
 &\tt{task5}& \tt{93} &{\tiny $\pm$\tt{13}} & \tt{\color{myblue}{95}} &{\tiny $\pm$\tt{6}} & \tt{\color{myblue}{100}} &{\tiny $\pm$\tt{0}} & \tt{0} &{\tiny $\pm$\tt{0}} & \tt{0} &{\tiny $\pm$\tt{0}} & \tt{0} &{\tiny $\pm$\tt{0}} & \tt{0} &{\tiny $\pm$\tt{0}} & \tt{0} &{\tiny $\pm$\tt{0}} & \tt{7} &{\tiny $\pm$\tt{9}} & \tt{\color{myblue}{100}} &{\tiny $\pm$\tt{0}} \\
 &\tt{overall}& \tt{\color{myblue}{98}} &{\tiny $\pm$\tt{3}} & \tt{91} &{\tiny $\pm$\tt{2}} & \tt{\color{myblue}{100}} &{\tiny $\pm$\tt{0}} & \tt{1} &{\tiny $\pm$\tt{1}} & \tt{1} &{\tiny $\pm$\tt{1}} & \tt{19} &{\tiny $\pm$\tt{2}} & \tt{15} &{\tiny $\pm$\tt{5}} & \tt{1} &{\tiny $\pm$\tt{1}} & \tt{12} &{\tiny $\pm$\tt{6}} & \tt{\color{myblue}{100}} &{\tiny $\pm$\tt{0}} \\
\midrule
\multirow[c]{6}{*}{\tt{puzzle-4x5-play-oraclerep-v0} }&\tt{task1}& \tt{\color{myblue}{98}} &{\tiny $\pm$\tt{3}} & \tt{73} &{\tiny $\pm$\tt{20}} & \tt{\color{myblue}{100}} &{\tiny $\pm$\tt{0}} & \tt{0} &{\tiny $\pm$\tt{0}} & \tt{0} &{\tiny $\pm$\tt{0}} & \tt{0} &{\tiny $\pm$\tt{0}} & \tt{0} &{\tiny $\pm$\tt{0}} & \tt{7} &{\tiny $\pm$\tt{13}} & \tt{0} &{\tiny $\pm$\tt{0}} & \tt{\color{myblue}{100}} &{\tiny $\pm$\tt{0}} \\
 &\tt{task2}& \tt{0} &{\tiny $\pm$\tt{0}} & \tt{15} &{\tiny $\pm$\tt{18}} & \tt{\color{myblue}{100}} &{\tiny $\pm$\tt{0}} & \tt{0} &{\tiny $\pm$\tt{0}} & \tt{0} &{\tiny $\pm$\tt{0}} & \tt{0} &{\tiny $\pm$\tt{0}} & \tt{0} &{\tiny $\pm$\tt{0}} & \tt{0} &{\tiny $\pm$\tt{0}} & \tt{0} &{\tiny $\pm$\tt{0}} & \tt{\color{myblue}{100}} &{\tiny $\pm$\tt{0}} \\
 &\tt{task3}& \tt{0} &{\tiny $\pm$\tt{0}} & \tt{0} &{\tiny $\pm$\tt{0}} & \tt{\color{myblue}{97}} &{\tiny $\pm$\tt{4}} & \tt{0} &{\tiny $\pm$\tt{0}} & \tt{0} &{\tiny $\pm$\tt{0}} & \tt{0} &{\tiny $\pm$\tt{0}} & \tt{0} &{\tiny $\pm$\tt{0}} & \tt{0} &{\tiny $\pm$\tt{0}} & \tt{0} &{\tiny $\pm$\tt{0}} & \tt{\color{myblue}{100}} &{\tiny $\pm$\tt{0}} \\
 &\tt{task4}& \tt{0} &{\tiny $\pm$\tt{0}} & \tt{8} &{\tiny $\pm$\tt{8}} & \tt{92} &{\tiny $\pm$\tt{6}} & \tt{0} &{\tiny $\pm$\tt{0}} & \tt{0} &{\tiny $\pm$\tt{0}} & \tt{0} &{\tiny $\pm$\tt{0}} & \tt{0} &{\tiny $\pm$\tt{0}} & \tt{0} &{\tiny $\pm$\tt{0}} & \tt{0} &{\tiny $\pm$\tt{0}} & \tt{\color{myblue}{100}} &{\tiny $\pm$\tt{0}} \\
 &\tt{task5}& \tt{0} &{\tiny $\pm$\tt{0}} & \tt{0} &{\tiny $\pm$\tt{0}} & \tt{68} &{\tiny $\pm$\tt{13}} & \tt{0} &{\tiny $\pm$\tt{0}} & \tt{0} &{\tiny $\pm$\tt{0}} & \tt{0} &{\tiny $\pm$\tt{0}} & \tt{0} &{\tiny $\pm$\tt{0}} & \tt{0} &{\tiny $\pm$\tt{0}} & \tt{0} &{\tiny $\pm$\tt{0}} & \tt{\color{myblue}{93}} &{\tiny $\pm$\tt{13}} \\
 &\tt{overall}& \tt{20} &{\tiny $\pm$\tt{1}} & \tt{19} &{\tiny $\pm$\tt{4}} & \tt{91} &{\tiny $\pm$\tt{4}} & \tt{0} &{\tiny $\pm$\tt{0}} & \tt{0} &{\tiny $\pm$\tt{0}} & \tt{0} &{\tiny $\pm$\tt{0}} & \tt{0} &{\tiny $\pm$\tt{0}} & \tt{1} &{\tiny $\pm$\tt{3}} & \tt{0} &{\tiny $\pm$\tt{0}} & \tt{\color{myblue}{99}} &{\tiny $\pm$\tt{3}} \\
\bottomrule
\end{tabularew}
}
\end{table}

\begin{table}[h!]
\caption{
\footnotesize \textbf{Complete results for reward-based experiments.}
}

\label{table:full_standard}
\centering
\scalebox{0.659}
{
\begin{tabularew}{ll*{9}{>{\spew{.5}{+1}}r@{\,}l}}
\toprule
& \multicolumn{1}{c}{} & \multicolumn{8}{c}{\tt{Model-Free}} & \multicolumn{10}{c}{\tt{Model-Based}} \\
\cmidrule(r){3-10}
\cmidrule(r){11-20}
\multicolumn{1}{l}{\tt{Environment}} & \multicolumn{1}{l}{\tt{Task}} & \multicolumn{2}{c}{\tt{IQL}} & \multicolumn{2}{c}{\tt{ReBRAC}} & \multicolumn{2}{c}{\tt{IDQL}} & \multicolumn{2}{c}{\tt{FQL}} & \multicolumn{2}{c}{\tt{MOPO}} & \multicolumn{2}{c}{\tt{MOBILE}} & \multicolumn{2}{c}{\tt{LEQ}} & \multicolumn{2}{c}{\tt{FMPC}} & \multicolumn{2}{c}{\tt{\color{myblue}\tt{MAC}}} \\
\midrule
\multirow[c]{6}{*}{\tt{cube-single-play-singletask-v0} }&\tt{task1}& \tt{88} &{\tiny $\pm$\tt{3}} & \tt{89} &{\tiny $\pm$\tt{5}} & \tt{\color{myblue}{95}} &{\tiny $\pm$\tt{2}} & \tt{\color{myblue}{97}} &{\tiny $\pm$\tt{2}} & \tt{12} &{\tiny $\pm$\tt{16}} & \tt{85} &{\tiny $\pm$\tt{22}} & \tt{0} &{\tiny $\pm$\tt{0}} & \tt{10} &{\tiny $\pm$\tt{9}} & \tt{\color{myblue}{100}} &{\tiny $\pm$\tt{0}} \\
 &\tt{task2}& \tt{85} &{\tiny $\pm$\tt{8}} & \tt{92} &{\tiny $\pm$\tt{4}} & \tt{\color{myblue}{96}} &{\tiny $\pm$\tt{2}} & \tt{\color{myblue}{97}} &{\tiny $\pm$\tt{2}} & \tt{10} &{\tiny $\pm$\tt{16}} & \tt{80} &{\tiny $\pm$\tt{12}} & \tt{0} &{\tiny $\pm$\tt{0}} & \tt{8} &{\tiny $\pm$\tt{8}} & \tt{\color{myblue}{100}} &{\tiny $\pm$\tt{0}} \\
 &\tt{task3}& \tt{91} &{\tiny $\pm$\tt{5}} & \tt{93} &{\tiny $\pm$\tt{3}} & \tt{\color{myblue}{99}} &{\tiny $\pm$\tt{1}} & \tt{\color{myblue}{98}} &{\tiny $\pm$\tt{2}} & \tt{15} &{\tiny $\pm$\tt{14}} & \tt{83} &{\tiny $\pm$\tt{17}} & \tt{0} &{\tiny $\pm$\tt{0}} & \tt{10} &{\tiny $\pm$\tt{9}} & \tt{\color{myblue}{98}} &{\tiny $\pm$\tt{3}} \\
 &\tt{task4}& \tt{73} &{\tiny $\pm$\tt{6}} & \tt{92} &{\tiny $\pm$\tt{3}} & \tt{93} &{\tiny $\pm$\tt{4}} & \tt{\color{myblue}{94}} &{\tiny $\pm$\tt{3}} & \tt{2} &{\tiny $\pm$\tt{3}} & \tt{72} &{\tiny $\pm$\tt{19}} & \tt{0} &{\tiny $\pm$\tt{0}} & \tt{13} &{\tiny $\pm$\tt{9}} & \tt{\color{myblue}{98}} &{\tiny $\pm$\tt{3}} \\
 &\tt{task5}& \tt{78} &{\tiny $\pm$\tt{9}} & \tt{87} &{\tiny $\pm$\tt{8}} & \tt{90} &{\tiny $\pm$\tt{6}} & \tt{\color{myblue}{93}} &{\tiny $\pm$\tt{3}} & \tt{20} &{\tiny $\pm$\tt{26}} & \tt{87} &{\tiny $\pm$\tt{19}} & \tt{0} &{\tiny $\pm$\tt{0}} & \tt{3} &{\tiny $\pm$\tt{4}} & \tt{\color{myblue}{97}} &{\tiny $\pm$\tt{7}} \\
 &\tt{overall}& \tt{83} &{\tiny $\pm$\tt{9}} & \tt{91} &{\tiny $\pm$\tt{5}} & \tt{\color{myblue}{95}} &{\tiny $\pm$\tt{4}} & \tt{\color{myblue}{96}} &{\tiny $\pm$\tt{3}} & \tt{12} &{\tiny $\pm$\tt{4}} & \tt{81} &{\tiny $\pm$\tt{8}} & \tt{0} &{\tiny $\pm$\tt{0}} & \tt{9} &{\tiny $\pm$\tt{5}} & \tt{\color{myblue}{99}} &{\tiny $\pm$\tt{2}} \\
\midrule
\multirow[c]{6}{*}{\tt{cube-double-play-singletask-v0} }&\tt{task1}& \tt{27} &{\tiny $\pm$\tt{5}} & \tt{45} &{\tiny $\pm$\tt{6}} & \tt{39} &{\tiny $\pm$\tt{19}} & \tt{61} &{\tiny $\pm$\tt{9}} & \tt{2} &{\tiny $\pm$\tt{3}} & \tt{7} &{\tiny $\pm$\tt{8}} & \tt{0} &{\tiny $\pm$\tt{0}} & \tt{15} &{\tiny $\pm$\tt{10}} & \tt{\color{myblue}{82}} &{\tiny $\pm$\tt{15}} \\
 &\tt{task2}& \tt{1} &{\tiny $\pm$\tt{1}} & \tt{7} &{\tiny $\pm$\tt{3}} & \tt{16} &{\tiny $\pm$\tt{10}} & \tt{36} &{\tiny $\pm$\tt{6}} & \tt{0} &{\tiny $\pm$\tt{0}} & \tt{0} &{\tiny $\pm$\tt{0}} & \tt{0} &{\tiny $\pm$\tt{0}} & \tt{0} &{\tiny $\pm$\tt{0}} & \tt{\color{myblue}{50}} &{\tiny $\pm$\tt{12}} \\
 &\tt{task3}& \tt{0} &{\tiny $\pm$\tt{0}} & \tt{4} &{\tiny $\pm$\tt{1}} & \tt{17} &{\tiny $\pm$\tt{8}} & \tt{22} &{\tiny $\pm$\tt{5}} & \tt{2} &{\tiny $\pm$\tt{3}} & \tt{0} &{\tiny $\pm$\tt{0}} & \tt{0} &{\tiny $\pm$\tt{0}} & \tt{0} &{\tiny $\pm$\tt{0}} & \tt{\color{myblue}{55}} &{\tiny $\pm$\tt{10}} \\
 &\tt{task4}& \tt{0} &{\tiny $\pm$\tt{0}} & \tt{1} &{\tiny $\pm$\tt{1}} & \tt{0} &{\tiny $\pm$\tt{1}} & \tt{5} &{\tiny $\pm$\tt{2}} & \tt{0} &{\tiny $\pm$\tt{0}} & \tt{0} &{\tiny $\pm$\tt{0}} & \tt{0} &{\tiny $\pm$\tt{0}} & \tt{0} &{\tiny $\pm$\tt{0}} & \tt{\color{myblue}{28}} &{\tiny $\pm$\tt{8}} \\
 &\tt{task5}& \tt{4} &{\tiny $\pm$\tt{3}} & \tt{4} &{\tiny $\pm$\tt{2}} & \tt{1} &{\tiny $\pm$\tt{1}} & \tt{19} &{\tiny $\pm$\tt{10}} & \tt{2} &{\tiny $\pm$\tt{3}} & \tt{0} &{\tiny $\pm$\tt{0}} & \tt{0} &{\tiny $\pm$\tt{0}} & \tt{0} &{\tiny $\pm$\tt{0}} & \tt{\color{myblue}{50}} &{\tiny $\pm$\tt{9}} \\
 &\tt{overall}& \tt{7} &{\tiny $\pm$\tt{11}} & \tt{12} &{\tiny $\pm$\tt{17}} & \tt{15} &{\tiny $\pm$\tt{17}} & \tt{29} &{\tiny $\pm$\tt{21}} & \tt{1} &{\tiny $\pm$\tt{1}} & \tt{1} &{\tiny $\pm$\tt{2}} & \tt{0} &{\tiny $\pm$\tt{0}} & \tt{3} &{\tiny $\pm$\tt{2}} & \tt{\color{myblue}{53}} &{\tiny $\pm$\tt{4}} \\
\midrule
\multirow[c]{6}{*}{\tt{scene-play-singletask-v0} }&\tt{task1}& \tt{94} &{\tiny $\pm$\tt{3}} & \tt{\color{myblue}{95}} &{\tiny $\pm$\tt{2}} & \tt{\color{myblue}{100}} &{\tiny $\pm$\tt{0}} & \tt{\color{myblue}{100}} &{\tiny $\pm$\tt{0}} & \tt{30} &{\tiny $\pm$\tt{38}} & \tt{37} &{\tiny $\pm$\tt{16}} & \tt{0} &{\tiny $\pm$\tt{0}} & \tt{15} &{\tiny $\pm$\tt{15}} & \tt{\color{myblue}{100}} &{\tiny $\pm$\tt{0}} \\
 &\tt{task2}& \tt{12} &{\tiny $\pm$\tt{3}} & \tt{50} &{\tiny $\pm$\tt{13}} & \tt{33} &{\tiny $\pm$\tt{14}} & \tt{76} &{\tiny $\pm$\tt{9}} & \tt{2} &{\tiny $\pm$\tt{3}} & \tt{5} &{\tiny $\pm$\tt{10}} & \tt{0} &{\tiny $\pm$\tt{0}} & \tt{3} &{\tiny $\pm$\tt{4}} & \tt{\color{myblue}{100}} &{\tiny $\pm$\tt{0}} \\
 &\tt{task3}& \tt{32} &{\tiny $\pm$\tt{7}} & \tt{55} &{\tiny $\pm$\tt{16}} & \tt{\color{myblue}{94}} &{\tiny $\pm$\tt{4}} & \tt{\color{myblue}{98}} &{\tiny $\pm$\tt{1}} & \tt{0} &{\tiny $\pm$\tt{0}} & \tt{0} &{\tiny $\pm$\tt{0}} & \tt{0} &{\tiny $\pm$\tt{0}} & \tt{0} &{\tiny $\pm$\tt{0}} & \tt{\color{myblue}{95}} &{\tiny $\pm$\tt{10}} \\
 &\tt{task4}& \tt{0} &{\tiny $\pm$\tt{1}} & \tt{3} &{\tiny $\pm$\tt{3}} & \tt{4} &{\tiny $\pm$\tt{3}} & \tt{5} &{\tiny $\pm$\tt{1}} & \tt{0} &{\tiny $\pm$\tt{0}} & \tt{0} &{\tiny $\pm$\tt{0}} & \tt{0} &{\tiny $\pm$\tt{0}} & \tt{0} &{\tiny $\pm$\tt{0}} & \tt{\color{myblue}{95}} &{\tiny $\pm$\tt{6}} \\
 &\tt{task5}& \tt{0} &{\tiny $\pm$\tt{0}} & \tt{0} &{\tiny $\pm$\tt{0}} & \tt{0} &{\tiny $\pm$\tt{0}} & \tt{0} &{\tiny $\pm$\tt{0}} & \tt{0} &{\tiny $\pm$\tt{0}} & \tt{0} &{\tiny $\pm$\tt{0}} & \tt{0} &{\tiny $\pm$\tt{0}} & \tt{2} &{\tiny $\pm$\tt{3}} & \tt{\color{myblue}{93}} &{\tiny $\pm$\tt{8}} \\
 &\tt{overall}& \tt{28} &{\tiny $\pm$\tt{36}} & \tt{41} &{\tiny $\pm$\tt{37}} & \tt{46} &{\tiny $\pm$\tt{44}} & \tt{56} &{\tiny $\pm$\tt{45}} & \tt{6} &{\tiny $\pm$\tt{8}} & \tt{8} &{\tiny $\pm$\tt{4}} & \tt{0} &{\tiny $\pm$\tt{0}} & \tt{4} &{\tiny $\pm$\tt{4}} & \tt{\color{myblue}{97}} &{\tiny $\pm$\tt{4}} \\
\midrule
\multirow[c]{6}{*}{\tt{puzzle-3x3-play-singletask-v0} }&\tt{task1}& \tt{33} &{\tiny $\pm$\tt{6}} & \tt{\color{myblue}{97}} &{\tiny $\pm$\tt{4}} & \tt{52} &{\tiny $\pm$\tt{12}} & \tt{90} &{\tiny $\pm$\tt{4}} & \tt{\color{myblue}{100}} &{\tiny $\pm$\tt{0}} & \tt{60} &{\tiny $\pm$\tt{47}} & \tt{52} &{\tiny $\pm$\tt{36}} & \tt{5} &{\tiny $\pm$\tt{3}} & \tt{\color{myblue}{100}} &{\tiny $\pm$\tt{0}} \\
 &\tt{task2}& \tt{4} &{\tiny $\pm$\tt{3}} & \tt{1} &{\tiny $\pm$\tt{1}} & \tt{0} &{\tiny $\pm$\tt{1}} & \tt{\color{myblue}{16}} &{\tiny $\pm$\tt{5}} & \tt{0} &{\tiny $\pm$\tt{0}} & \tt{0} &{\tiny $\pm$\tt{0}} & \tt{0} &{\tiny $\pm$\tt{0}} & \tt{0} &{\tiny $\pm$\tt{0}} & \tt{0} &{\tiny $\pm$\tt{0}} \\
 &\tt{task3}& \tt{3} &{\tiny $\pm$\tt{2}} & \tt{3} &{\tiny $\pm$\tt{1}} & \tt{0} &{\tiny $\pm$\tt{0}} & \tt{\color{myblue}{10}} &{\tiny $\pm$\tt{3}} & \tt{0} &{\tiny $\pm$\tt{0}} & \tt{0} &{\tiny $\pm$\tt{0}} & \tt{0} &{\tiny $\pm$\tt{0}} & \tt{0} &{\tiny $\pm$\tt{0}} & \tt{0} &{\tiny $\pm$\tt{0}} \\
 &\tt{task4}& \tt{2} &{\tiny $\pm$\tt{1}} & \tt{2} &{\tiny $\pm$\tt{1}} & \tt{0} &{\tiny $\pm$\tt{0}} & \tt{\color{myblue}{16}} &{\tiny $\pm$\tt{5}} & \tt{0} &{\tiny $\pm$\tt{0}} & \tt{0} &{\tiny $\pm$\tt{0}} & \tt{0} &{\tiny $\pm$\tt{0}} & \tt{0} &{\tiny $\pm$\tt{0}} & \tt{0} &{\tiny $\pm$\tt{0}} \\
 &\tt{task5}& \tt{3} &{\tiny $\pm$\tt{2}} & \tt{5} &{\tiny $\pm$\tt{3}} & \tt{0} &{\tiny $\pm$\tt{0}} & \tt{\color{myblue}{16}} &{\tiny $\pm$\tt{3}} & \tt{0} &{\tiny $\pm$\tt{0}} & \tt{0} &{\tiny $\pm$\tt{0}} & \tt{0} &{\tiny $\pm$\tt{0}} & \tt{2} &{\tiny $\pm$\tt{3}} & \tt{0} &{\tiny $\pm$\tt{0}} \\
 &\tt{overall}& \tt{9} &{\tiny $\pm$\tt{13}} & \tt{21} &{\tiny $\pm$\tt{38}} & \tt{10} &{\tiny $\pm$\tt{21}} & \tt{\color{myblue}{30}} &{\tiny $\pm$\tt{31}} & \tt{20} &{\tiny $\pm$\tt{0}} & \tt{12} &{\tiny $\pm$\tt{9}} & \tt{10} &{\tiny $\pm$\tt{7}} & \tt{1} &{\tiny $\pm$\tt{1}} & \tt{20} &{\tiny $\pm$\tt{0}} \\
\midrule
\multirow[c]{6}{*}{\tt{puzzle-4x4-play-singletask-v0} }&\tt{task1}& \tt{12} &{\tiny $\pm$\tt{2}} & \tt{26} &{\tiny $\pm$\tt{4}} & \tt{48} &{\tiny $\pm$\tt{5}} & \tt{34} &{\tiny $\pm$\tt{8}} & \tt{0} &{\tiny $\pm$\tt{0}} & \tt{0} &{\tiny $\pm$\tt{0}} & \tt{0} &{\tiny $\pm$\tt{0}} & \tt{0} &{\tiny $\pm$\tt{0}} & \tt{\color{myblue}{98}} &{\tiny $\pm$\tt{3}} \\
 &\tt{task2}& \tt{7} &{\tiny $\pm$\tt{4}} & \tt{12} &{\tiny $\pm$\tt{4}} & \tt{14} &{\tiny $\pm$\tt{5}} & \tt{16} &{\tiny $\pm$\tt{5}} & \tt{0} &{\tiny $\pm$\tt{0}} & \tt{0} &{\tiny $\pm$\tt{0}} & \tt{0} &{\tiny $\pm$\tt{0}} & \tt{0} &{\tiny $\pm$\tt{0}} & \tt{\color{myblue}{33}} &{\tiny $\pm$\tt{27}} \\
 &\tt{task3}& \tt{9} &{\tiny $\pm$\tt{3}} & \tt{15} &{\tiny $\pm$\tt{3}} & \tt{34} &{\tiny $\pm$\tt{5}} & \tt{18} &{\tiny $\pm$\tt{5}} & \tt{0} &{\tiny $\pm$\tt{0}} & \tt{0} &{\tiny $\pm$\tt{0}} & \tt{0} &{\tiny $\pm$\tt{0}} & \tt{0} &{\tiny $\pm$\tt{0}} & \tt{\color{myblue}{100}} &{\tiny $\pm$\tt{0}} \\
 &\tt{task4}& \tt{5} &{\tiny $\pm$\tt{2}} & \tt{10} &{\tiny $\pm$\tt{3}} & \tt{26} &{\tiny $\pm$\tt{6}} & \tt{11} &{\tiny $\pm$\tt{3}} & \tt{0} &{\tiny $\pm$\tt{0}} & \tt{0} &{\tiny $\pm$\tt{0}} & \tt{0} &{\tiny $\pm$\tt{0}} & \tt{0} &{\tiny $\pm$\tt{0}} & \tt{\color{myblue}{85}} &{\tiny $\pm$\tt{14}} \\
 &\tt{task5}& \tt{4} &{\tiny $\pm$\tt{1}} & \tt{7} &{\tiny $\pm$\tt{3}} & \tt{24} &{\tiny $\pm$\tt{11}} & \tt{7} &{\tiny $\pm$\tt{3}} & \tt{0} &{\tiny $\pm$\tt{0}} & \tt{0} &{\tiny $\pm$\tt{0}} & \tt{0} &{\tiny $\pm$\tt{0}} & \tt{0} &{\tiny $\pm$\tt{0}} & \tt{\color{myblue}{72}} &{\tiny $\pm$\tt{40}} \\
 &\tt{overall}& \tt{7} &{\tiny $\pm$\tt{4}} & \tt{14} &{\tiny $\pm$\tt{8}} & \tt{29} &{\tiny $\pm$\tt{13}} & \tt{17} &{\tiny $\pm$\tt{10}} & \tt{0} &{\tiny $\pm$\tt{0}} & \tt{0} &{\tiny $\pm$\tt{0}} & \tt{0} &{\tiny $\pm$\tt{0}} & \tt{0} &{\tiny $\pm$\tt{0}} & \tt{\color{myblue}{78}} &{\tiny $\pm$\tt{13}} \\
\bottomrule
\end{tabularew}
}
\end{table}

\clearpage
\section{Training Curves}
\begin{figure}[h!]
    \centering
    \includegraphics[width=0.87\textwidth]{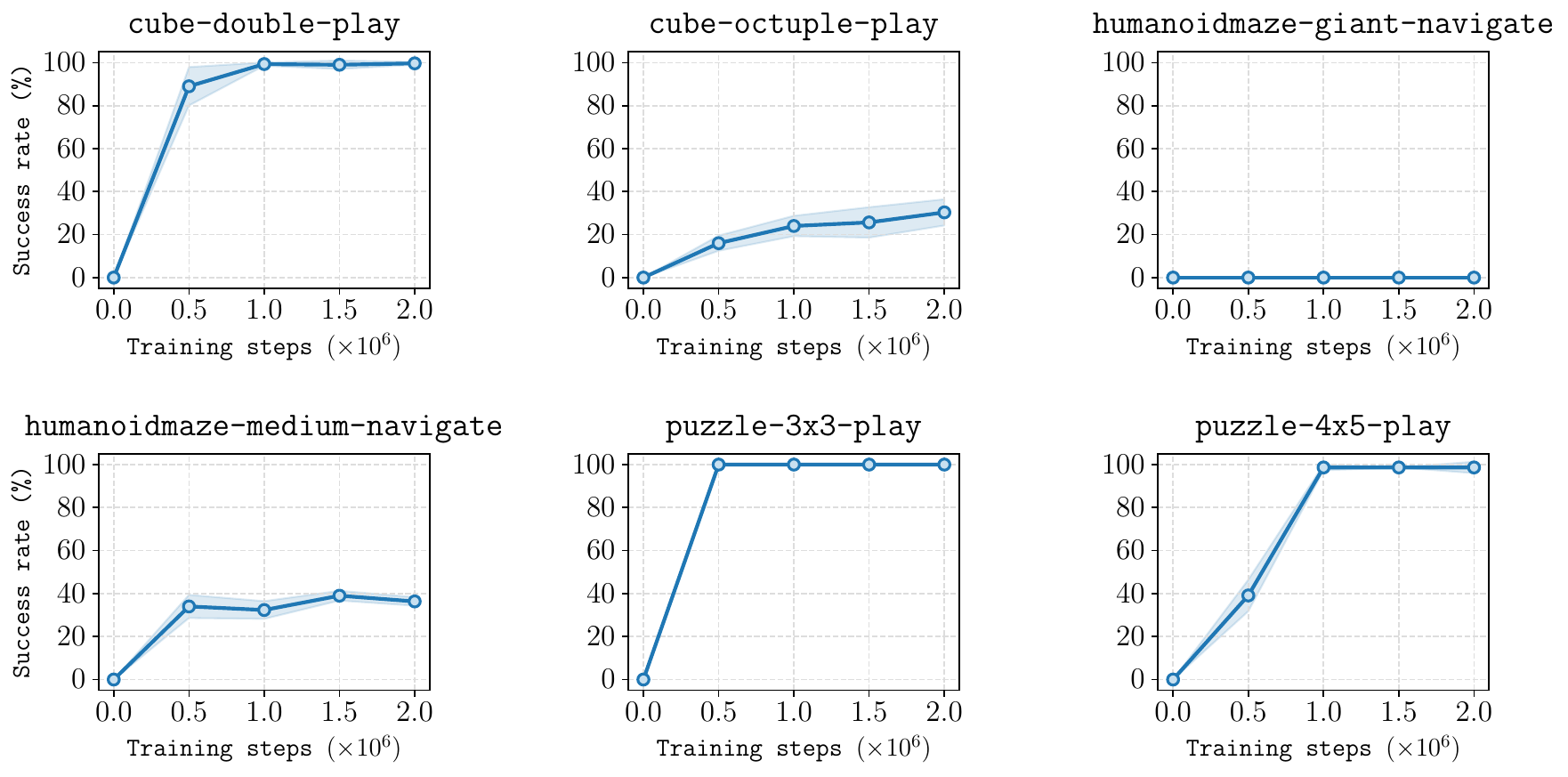}
    \caption{
    \footnotesize
    \textbf{Training curve of \methodname in large-scale, long-horizon environments.} We report the success rate for 15 evaluation episodes across 4 seeds (total 60 episodes). Shaded region represents the [mean - std, mean + std].
    }
    \label{fig:training_curves_multitask}
\end{figure}

\begin{figure}[h!]
    \centering
    \includegraphics[width=1.0\textwidth]{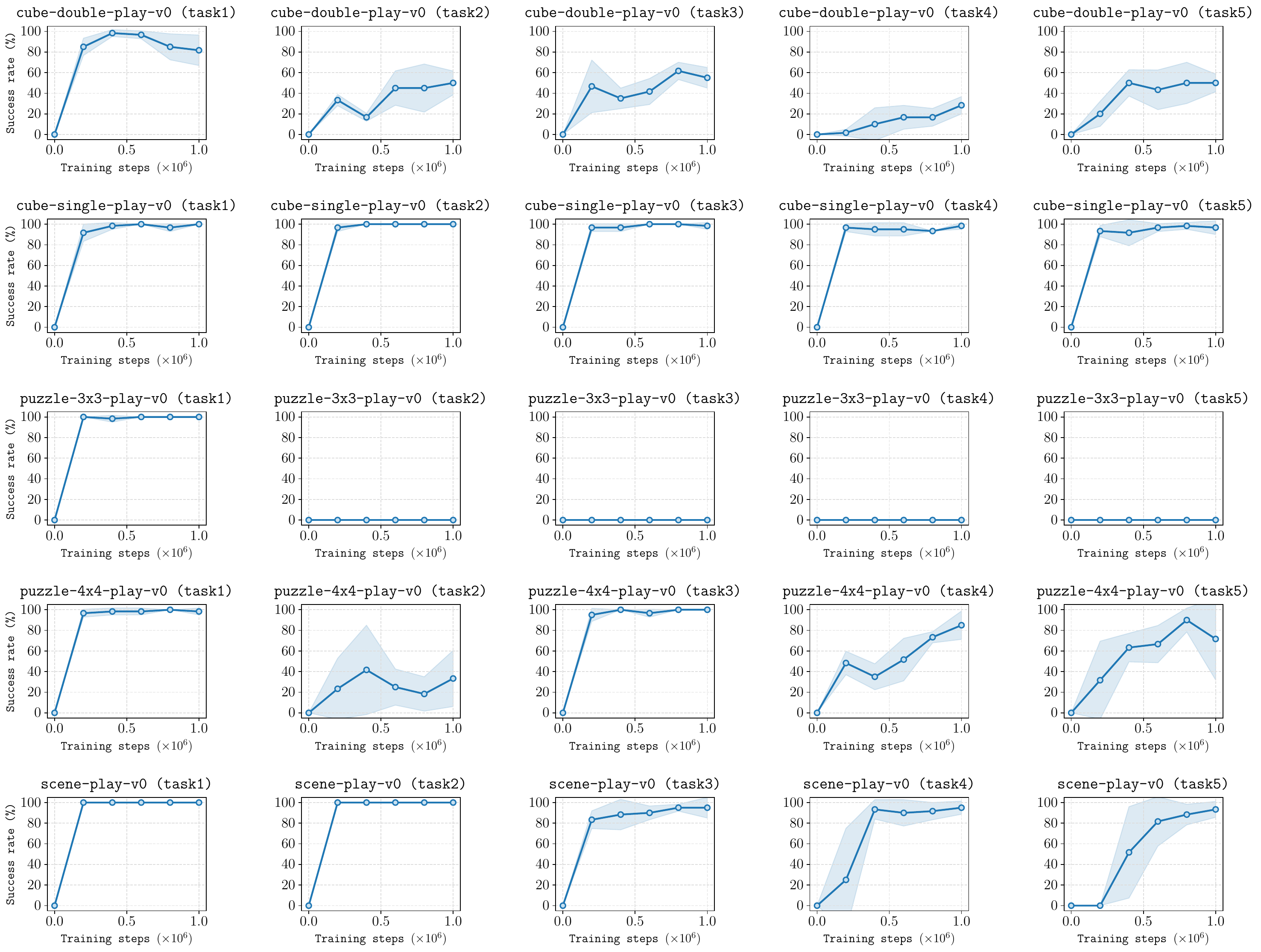}
    \caption{
    \footnotesize
    \textbf{Training curve of \methodname in reward-based environments.} We report the success rate for 15 evaluation episodes across 4 seeds (total 60 episodes). Shaded region represents the [mean - std, mean + std].}
    
    \label{fig:training_curves_singletask}
\end{figure}

We provide the training curve of \methodname for large-scale, long-horizon environments and reward-based environment in \Cref{fig:training_curves_multitask} and \Cref{fig:training_curves_singletask} (corresponding to \Cref{table:long_100m} and \Cref{table:standard}). We plot the mean and the standard deviation (across 4 seeds) by  covering [mean - std, mean + std] area with a lighter color.

\clearpage
\section{More Ablation Experiments}
\ul{\textbf{Q: Is distillation with $\pi_\theta$ necessary?}}

\begin{wraptable}{r}{0.48\linewidth}
    \vspace{-1.3em}
    \centering
    \caption{
    \footnotesize
    \textbf{Ablation of using $\pi_\theta$.}
    }
    \vspace{-1em}
    \vspace{5pt}
    \label{table:abl2}
    \raisebox{0pt}[\dimexpr\height-1.0\baselineskip\relax]{
    \scalebox{0.67}{

\begin{tabularew}{l*{3}{>{\spew{.5}{+1}}r@{\,}l}}
\toprule
\multicolumn{1}{l}{\tt{Task}} & \multicolumn{2}{c}{\tt{\methodname (w/o $\pi_\theta$)}} & \multicolumn{2}{c}{\tt{\color{myblue}\tt{\methodname}}} \\
\midrule
\tt{cube-single-play-v0} & \tt{2} &{\tiny $\pm$\tt{3}} & \tt{\color{myblue}{100}} &{\tiny $\pm$\tt{0}} \\
\tt{cube-double-play-v0} & \tt{0} &{\tiny $\pm$\tt{0}} & \tt{\color{myblue}{50}} &{\tiny $\pm$\tt{12}} \\
\tt{scene-play-v0} & \tt{5} &{\tiny $\pm$\tt{9}} & \tt{\color{myblue}{100}} &{\tiny $\pm$\tt{0}} \\
\tt{puzzle-3x3-play-v0} & \tt{\color{myblue}{2}} &{\tiny $\pm$\tt{3}} & \tt{{0}} &{\tiny $\pm$\tt{0}} \\
\tt{puzzle-4x4-play-v0} & \tt{15} &{\tiny $\pm$\tt{6}} & \tt{\color{myblue}{85}} &{\tiny $\pm$\tt{14}} \\
\bottomrule
\end{tabularew}

    }
    }
    \vspace{-0.5em}
\end{wraptable}
\textbf{A:} 
To understand the importance of this component, we conduct an ablation study of \methodname removing the distillation. Specifically, we directly train the one-step flow model $\pi_\omega$ with flow matching BC loss, instead of distilling the multi-step flow model $\pi_\theta$. 
We present the ablation results on the default tasks for five reward-based environments used in \Cref{table:abl2}.
The results indicate that the use of $\pi_\theta$ is crucial for \methodname in OGBench tasks where behavioral policies are highly multi-modal.

\ul{\textbf{Q: How does model error correlate with the performance?}}

\begin{wrapfigure}{r}{0.5\textwidth}
    \vspace{-0.1em}
    \centering
    \raisebox{0pt}[\dimexpr\height-1.0\baselineskip\relax]{
        \begin{subfigure}[t]{1.0\linewidth}
        \includegraphics[width=0.383\linewidth]{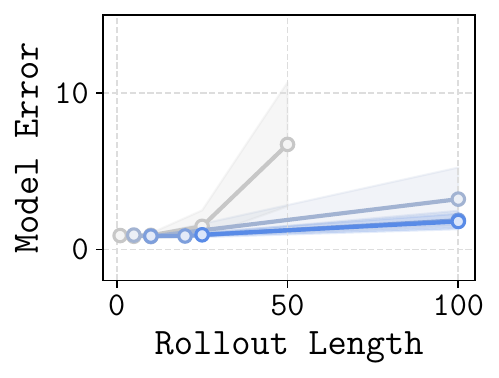}
        \includegraphics[width=0.55\linewidth]{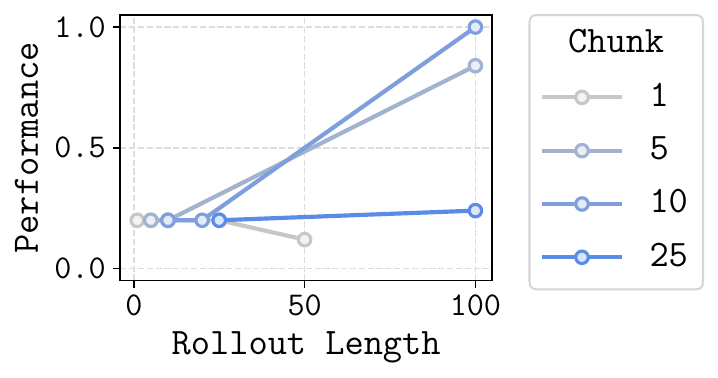}
        \end{subfigure}
    }
    \vspace{-2.0em}
    \caption{
    \footnotesize
        \textbf{Model error correlate with the performance, unless action chunk size is too large.}
    }
    \label{fig:mse_a1}
    \vspace{-1em}
\end{wrapfigure}

\textbf{A:} 
\Cref{fig:mse_a1} shows the policy performance and rollout error with respect to the rollout length for various action chunk sizes. For chunk sizes of 1, 5, and 10, we observe a consistent trend that at a given rollout length, smaller chunks produce larger model-prediction errors and correspondingly lower policy performance (clearly shown in rollout length of 50 and 100). However, excessively large chunk sizes (25) breaks this trend, where it achieves lower rollout error, but does not yield better performance. It is because while larger chunks helps reducing the compounding model error, they also make both policy learning and Q-function estimation harder since the action space grows exponentially. While flow rejection sampling mitigates this problem by limiting the action sequence to in-distribution actions, extremely large chunk sizes exacerbate this problem, hurting the performance despite improved model accuracy.

\end{document}